\pdfoutput=1

\documentclass[11pt]{article}

\usepackage[final]{acl}

\usepackage{times}
\usepackage{latexsym}

\usepackage[T1]{fontenc}

\usepackage[utf8]{inputenc}

\usepackage{microtype}

\usepackage{inconsolata}

\usepackage{CJKutf8}
\usepackage[normalem]{ulem}

\usepackage[inline]{enumitem}
\setlist{nosep}

\usepackage{tikz}
\usepackage{tikz-dependency}
\usepackage{amsmath}
\usepackage{graphicx}
\usepackage{pgfplots}
\pgfplotsset{compat=1.8}
\usepackage{subcaption}
\usepackage{tikz-qtree}
\usepgflibrary{arrows.meta}
\usepgfplotslibrary{fillbetween}
\usetikzlibrary{fit, calc, decorations.pathreplacing, positioning, shapes.geometric, backgrounds}
\usepackage{float}
\usepackage{mathpartir}

\usepackage[commentColor=blue]{algpseudocodex}
\usepackage{algorithm}
\usepackage{algorithmicx}

\usepackage{booktabs}
\usepackage{multirow}

\usepackage{amssymb}

\usepackage{hyperref}

\def \xmidrule[#1]{%
\noalign{\vskip\aboverulesep}%
\multispan{#1}{\leaders\hbox to 4pt{\hss\vrule\@height\cmidrulewidth\@width 2pt \relax\hss}\hfill\kern0pt}\cr%
\noalign{\vskip\belowrulesep}%
}

\DeclareDocumentCommand{\trapezoid}{O{2.0} O{1.0} O{0.5} m m m}{
  \begin{scope}[scale=0.9,thick]
    \draw[anchor=mid] (0, 0) -- (0, -{#2}) node[below=9pt,anchor=base] {\footnotesize\ensuremath{#4}} -- ({#1}, -{#2}) node [below=9pt,anchor=base] {\footnotesize\ensuremath{#5}} -- ({#1}, -{#3}) -- cycle;
    \draw node[] at ($(0, 0)!0.5!({#1}, {#3}) + (0, 0.2)$) {\footnotesize\ensuremath{#6}};
  \end{scope}
}

\DeclareDocumentCommand{\trapezoidarc}{O{2.0} O{1.0} O{0.5} m m m}{
  \begin{scope}[scale=0.9,thick]
    \draw[anchor=mid] (0, 0) -- (0, -{#2}) node[below=9pt,anchor=base] {\footnotesize\ensuremath{#4}} -- ({#1}, -{#2}) node [below=9pt,anchor=base] {\footnotesize\ensuremath{#5}} -- ({#1}, -{#3}) -- cycle;
    \draw node[] at ($(0, 0)!0.5!({#1}, {#3}) + (0, 0.2)$) {\footnotesize\ensuremath{#6}};
    \draw[->] (0, -{#2}) to[bend left=60] ({#1}, -{#2});
  \end{scope}
}

\DeclareDocumentCommand{\trapezoidarcdash}{O{2.0} O{1.0} O{0.5} m m m}{
  \begin{scope}[scale=0.9,thick]
    \draw[anchor=mid,densely dashed] (0, 0) -- (0, -{#2}) node[below=9pt,anchor=base] {\footnotesize\ensuremath{#4}} -- ({#1}, -{#2}) node [below=9pt,anchor=base] {\footnotesize\ensuremath{#5}} -- ({#1}, -{#3}) -- cycle;
    \draw node[] at ($(0, 0)!0.5!({#1}, {#3}) + (0, 0.2)$) {\footnotesize\ensuremath{#6}};
    \draw[->,densely dashed] (0, -{#2}) to[bend left=60] ({#1}, -{#2});
  \end{scope}
}

\DeclareDocumentCommand{\square}{O{0.5} O{0.5} m m m}{
  \begin{scope}[scale=0.9,thick]
    \draw[anchor=mid] (0, -{#2}) node[below left=9pt and 3pt,anchor=base] {\footnotesize\ensuremath{#3}} -- ({#1}, -{#2}) node (square) [below=9pt,anchor=base] {\footnotesize\ensuremath{#4}} -- ({#1}, 0) --  (0, 0) -- cycle;
    \draw node[] at ($(0, 0)!0.5!({#1}, 0) + (0, 0.2)$) {\footnotesize\ensuremath{#5}};
  \end{scope}
}

\DeclareDocumentCommand{\lefttriangle}{O{0.5} O{0.5} m m m}{
  \begin{scope}[scale=0.9,thick]
    \draw[anchor=mid] (0, -{#2}) node[below left=9pt and 3pt,anchor=base] {\footnotesize\ensuremath{#3}} -- ({#1}, -{#2}) node [below=9pt,anchor=base] {\footnotesize\ensuremath{#4}} -- ({#1}, 0) -- cycle;
    \draw node[] at ($(0, -{#2})!0.5!({#1}, 0) + (0, 0.45)$) {\footnotesize\ensuremath{#5}};
  \end{scope}
}

\DeclareDocumentCommand{\lefttriangledash}{O{0.5} O{0.5} m m m}{
  \begin{scope}[scale=0.9,thick]
    \draw[anchor=mid, densely dashed] (0, -{#2}) node[below left=9pt and 3pt,anchor=base] {\footnotesize\ensuremath{#3}} -- ({#1}, -{#2}) node [below=9pt,anchor=base] {\footnotesize\ensuremath{#4}} -- ({#1}, 0) -- cycle;
    \draw node[densely dashed] at ($(0, -{#2})!0.5!({#1}, 0) + (0, 0.45)$) {\footnotesize\ensuremath{#5}};
  \end{scope}
}

\DeclareDocumentCommand{\righttriangle}{O{0.5} O{0.5} m m m}{
  \begin{scope}[scale=0.9,thick]
    \draw[anchor=mid](0, 0) -- (0, -{#2}) node[below=9pt,anchor=base] {\footnotesize\ensuremath{#3}} -- ({#1}, -{#2}) node [below right=9pt and 3pt,anchor=base] {\footnotesize\ensuremath{#4}} -- cycle;
    \draw node[] at ($(0, 0)!0.5!({#1}, -{#2}) + (0, 0.45)$) {\footnotesize\ensuremath{#5}};
  \end{scope}
}

\DeclareDocumentCommand{\righttriangledash}{O{0.5} O{0.5} m m m}{
  \begin{scope}[scale=0.9,thick]
    \draw[anchor=mid, densely dashed](0, 0) -- (0, -{#2}) node[below=9pt,anchor=base] {\footnotesize\ensuremath{#3}} -- ({#1}, -{#2}) node [below right=9pt and 3pt,anchor=base] {\footnotesize\ensuremath{#4}} -- cycle;
    \draw node[densely dashed] at ($(0, 0)!0.5!({#1}, -{#2}) + (0, 0.45)$) {\footnotesize\ensuremath{#5}};
  \end{scope}
}

\DeclareDocumentCommand{\structra}{}{
  \begin{scope}[scale=0.9, >=stealth]
    \node[circle, fill, black, inner sep=1pt] (a) at (0, 0) {};
    \node[circle, draw, black, inner sep=1pt] (b) at (0.5, 0) {};
    \draw[->, out=60, in=120, rounded corners=0.1] (a.north) to  (b.north);
  \end{scope}
}

\DeclareDocumentCommand{\structbr}{}{
  \begin{scope}[scale=0.9, >=stealth]
    \node[circle, draw, black, inner sep=1pt] (a) at (0, 0) {};
    \node[circle, fill, black, inner sep=1pt] (b) at (0.5, 0) {};
    \draw[<-, out=60, in=120, rounded corners=0.1] (a.north) to  (b.north);
  \end{scope}
}

\DeclareDocumentCommand{\structrab}{}{
  \begin{scope}[scale=0.9, >=stealth]
    \node[circle, fill, black, inner sep=1pt] (a) at (0, 0) {};
    \node[circle, draw, black, inner sep=1pt] (b) at (0.5, 0) {};
    \node[circle, draw, black, inner sep=1pt] (c) at (1, 0) {};
    \draw[->, out=60, in=120, rounded corners=0.1] (a.north) to  (b.north);
    \draw[->, out=60, in=120, rounded corners=0.1] (b.north) to  (c.north);
  \end{scope}
}

\DeclareDocumentCommand{\structbcr}{}{
  \begin{scope}[scale=0.9, >=stealth]
    \node[circle, draw, black, inner sep=1pt] (a) at (0, 0) {};
    \node[circle, draw, black, inner sep=1pt] (b) at (0.5, 0) {};
    \node[circle, fill, black, inner sep=1pt] (c) at (1, 0) {};
    \draw[<-, out=60, in=120] (a.north) to  (b.north);
    \draw[<-, out=60, in=120] (b.north) to  (c.north);
  \end{scope}
}

\DeclareDocumentCommand{\structbrb}{}{
  \begin{scope}[scale=0.9, >=stealth]
    \node[circle, draw, black, inner sep=1pt] (a) at (0, 0) {};
    \node[circle, fill, black, inner sep=1pt] (b) at (0.5, 0) {};
    \node[circle, draw, black, inner sep=1pt] (c) at (1, 0) {};
    \draw[<-, out=60, in=120] (a.north) to  (b.north);
    \draw[->, out=60, in=120] (b.north) to  (c.north);
  \end{scope}
}

\DeclareDocumentCommand{\structraa}{}{
  \begin{scope}[scale=0.9, >=stealth]
    \node[circle, fill, black, inner sep=1pt] (a) at (0, 0) {};
    \node[circle, draw, black, inner sep=1pt] (b) at (0.5, 0) {};
    \node[circle, draw, black, inner sep=1pt] (c) at (1, 0) {};
    \draw[->, out=45, in=135] (a.north) to  (b.north);
    \draw[->, out=45, in=135] (a.north) to  (c.north);
  \end{scope}
}

\DeclareDocumentCommand{\structccr}{}{
  \begin{scope}[scale=0.9, >=stealth]
    \node[circle, draw, black, inner sep=1pt] (a) at (0, 0) {};
    \node[circle, draw, black, inner sep=1pt] (b) at (0.5, 0) {};
    \node[circle, fill, black, inner sep=1pt] (c) at (1, 0) {};
    \draw[<-, out=45, in=135] (b.north) to  (c.north);
    \draw[<-, out=45, in=135] (a.north) to  (c.north);
  \end{scope}
}

\DeclareDocumentCommand{\structcar}{}{
  \begin{scope}[scale=0.9, >=stealth]
    \node[circle, draw, black, inner sep=1pt] (a) at (0, 0) {};
    \node[circle, draw, black, inner sep=1pt] (b) at (0.5, 0) {};
    \node[circle, fill, black, inner sep=1pt] (c) at (1, 0) {};
    \draw[->, out=45, in=135] ($(a.north) + (1pt, 0)$) to  (b.north);
    \draw[<-, out=45, in=135] ($(a.north) + (-1pt, 0)$) to  (c.north);
  \end{scope}
}

\DeclareDocumentCommand{\structbcdr}{}{
  \begin{scope}[scale=0.9, >=stealth]
    \node[circle, draw, black, inner sep=1pt] (a) at (0, 0) {};
    \node[circle, draw, black, inner sep=1pt] (b) at (0.5, 0) {};
    \node[circle, draw, black, inner sep=1pt] (c) at (1, 0) {};
    \node[circle, fill, black, inner sep=1pt] (d) at (1.5, 0) {};
    \draw[<-, out=60, in=120] (a.north) to  (b.north);
    \draw[<-, out=60, in=120] (b.north) to  (c.north);
    \draw[<-, out=60, in=120] (c.north) to  (d.north);
  \end{scope}
}

\DeclareDocumentCommand{\structraaa}{}{
  \begin{scope}[scale=0.9, >=stealth]
    \node[circle, fill, black, inner sep=1pt] (a) at (0, 0) {};
    \node[circle, draw, black, inner sep=1pt] (b) at (0.5, 0) {};
    \node[circle, draw, black, inner sep=1pt] (c) at (1, 0) {};
    \node[circle, draw, black, inner sep=1pt] (d) at (1.5, 0) {};
    \draw[->, out=30, in=150] (a.north) to  (b.north);
    \draw[->, out=30, in=150] (a.north) to  (c.north);
    \draw[->, out=30, in=150] (a.north) to  (d.north);
  \end{scope}
}

\DeclareDocumentCommand{\structdddr}{}{
  \begin{scope}[scale=0.9, >=stealth]
    \node[circle, draw, black, inner sep=1pt] (a) at (0, 0) {};
    \node[circle, draw, black, inner sep=1pt] (b) at (0.5, 0) {};
    \node[circle, draw, black, inner sep=1pt] (c) at (1, 0) {};
    \node[circle, fill, black, inner sep=1pt] (d) at (1.5, 0) {};
    \draw[<-, out=30, in=150] (a.north) to  (d.north);
    \draw[<-, out=30, in=150] (b.north) to  (d.north);
    \draw[<-, out=30, in=150] (c.north) to  (d.north);
  \end{scope}
}

\DeclareDocumentCommand{\structbddr}{}{
  \begin{scope}[scale=0.9, >=stealth]
    \node[circle, draw, black, inner sep=1pt] (a) at (0, 0) {};
    \node[circle, draw, black, inner sep=1pt] (b) at (0.5, 0) {};
    \node[circle, draw, black, inner sep=1pt] (c) at (1, 0) {};
    \node[circle, fill, black, inner sep=1pt] (d) at (1.5, 0) {};
    \draw[<-, out=45, in=135] (a.north) to  (b.north);
    \draw[<-, out=45, in=135] (b.north) to  (d.north);
    \draw[<-, out=45, in=135] (c.north) to  (d.north);
  \end{scope}
}

\DeclareDocumentCommand{\structbrdb}{}{
  \begin{scope}[scale=0.9, >=stealth]
    \node[circle, draw, black, inner sep=1pt] (a) at (0, 0) {};
    \node[circle, fill, black, inner sep=1pt] (b) at (0.5, 0) {};
    \node[circle, draw, black, inner sep=1pt] (c) at (1, 0) {};
    \node[circle, draw, black, inner sep=1pt] (d) at (1.5, 0) {};
    \draw[<-, out=45, in=135] (a.north) to  (b.north);
    \draw[->, out=45, in=135] (b.north) to  ($(d.north) + (1pt, 0)$);
    \draw[<-, out=45, in=135] (c.north) to  ($(d.north) + (-1pt, 0)$);
  \end{scope}
}

\DeclareDocumentCommand{\structraac}{}{
  \begin{scope}[scale=0.9, >=stealth]
    \node[circle, fill, black, inner sep=1pt] (a) at (0, 0) {};
    \node[circle, draw, black, inner sep=1pt] (b) at (0.5, 0) {};
    \node[circle, draw, black, inner sep=1pt] (c) at (1, 0) {};
    \node[circle, draw, black, inner sep=1pt] (d) at (1.5, 0) {};
    \draw[->, out=45, in=135] (a.north) to  (b.north);
    \draw[->, out=45, in=135] (a.north) to  (c.north);
    \draw[->, out=45, in=135] (c.north) to  (d.north);
  \end{scope}
}

\DeclareDocumentCommand{\structcarc}{}{
  \begin{scope}[scale=0.9, >=stealth]
    \node[circle, draw, black, inner sep=1pt] (a) at (0, 0) {};
    \node[circle, draw, black, inner sep=1pt] (b) at (0.5, 0) {};
    \node[circle, fill, black, inner sep=1pt] (c) at (1, 0) {};
    \node[circle, draw, black, inner sep=1pt] (d) at (1.5, 0) {};
    \draw[<-, out=45, in=135] ($(a.north) + (-1pt, 0)$) to  (c.north);
    \draw[->, out=45, in=135] ($(a.north) + (1pt, 0)$) to  (b.north);
    \draw[->, out=45, in=135] (c.north) to  (d.north);
  \end{scope}
}

\DeclareDocumentCommand{\structbcrc}{}{
  \begin{scope}[scale=0.9, >=stealth]
    \node[circle, draw, black, inner sep=1pt] (a) at (0, 0) {};
    \node[circle, draw, black, inner sep=1pt] (b) at (0.5, 0) {};
    \node[circle, fill, black, inner sep=1pt] (c) at (1, 0) {};
    \node[circle, draw, black, inner sep=1pt] (d) at (1.5, 0) {};
    \draw[<-, out=60, in=120] (a.north) to  (b.north);
    \draw[<-, out=60, in=120] (b.north) to  (c.north);
    \draw[->, out=60, in=120] (c.north) to  (d.north);
  \end{scope}
}

\title{Character-Level Chinese Dependency Parsing via Modeling Latent Intra-Word Structure}

\author{Yang Hou \and Zhenghua Li\thanks{$~$Corresponding author} \\
       School of Computer Science and Technology,
        \\Soochow University, China \\ \texttt{yhou1@stu.suda.edu.cn}$~~~~~$ 
       \texttt{zhli13@suda.edu.cn}}

\begin{document}
\begin{CJK}{UTF8}{gbsn}
\maketitle
\begin{abstract}
   Revealing the syntactic structure of sentences in Chinese poses significant challenges for word-level parsers due to the absence of clear word boundaries. To facilitate a transition from word-level to character-level Chinese dependency parsing, this paper proposes modeling latent internal structures within words. In this way, each word-level dependency tree is interpreted as a forest of character-level trees.
    A constrained Eisner algorithm is implemented to ensure the compatibility of character-level trees, guaranteeing a single root for intra-word structures and establishing inter-word dependencies between these roots. 
    Experiments on Chinese treebanks demonstrate the superiority of our method over both the pipeline framework and previous joint models.
    A detailed analysis reveals that a coarse-to-fine parsing strategy empowers the model to predict more linguistically plausible intra-word structures.
\end{abstract}

\section{Introduction}
\begin{figure*}[tb]
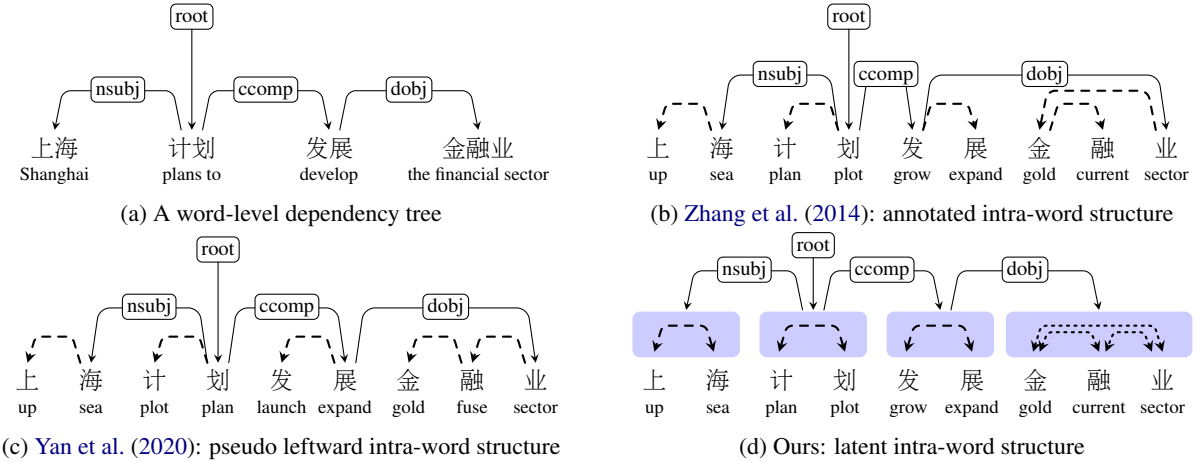

    \centering
    \tikzset{intra word arc/.style={dashed, thick}}
    \tikzset{latent intra word arc/.style={dotted, thick}}
    \tikzset{inter word arc/.style={}}
    \tikzstyle{HAN_WORD}=[draw=none, outer sep=0ex]
    \tikzstyle{EN_WORD}=[draw=none, inner sep=0pt,  text height=1ex, text depth=.5ex, font=\scriptsize]
    \tikzstyle{EN_CHAR}=[draw=none, inner sep=0pt, text height=1ex, text depth=.5ex, font=\scriptsize]
    \tikzstyle{WORD}=[draw=none, rounded corners=1mm, inner sep=0pt]
    \begin{subfigure}[b]{1\columnwidth}
        \centering
        \begin{dependency}[]
            \begin{deptext}[column sep=1cm, font=\small]
                上海 \& 计划 \& 发展 \& 金融业  \\
            \end{deptext}
            \deproot[edge height=10ex]{2}{root}
            \depedge[edge height=3ex]{2}{1}{nsubj}
            \depedge[edge height=3ex]{2}{3}{ccomp}
            \depedge[edge height=3ex]{3}{4}{dobj}

            \wordgroup[HAN_WORD]{1}{1}{1}{han_w1}
            \wordgroup[HAN_WORD]{1}{2}{2}{han_w2}
            \wordgroup[HAN_WORD]{1}{3}{3}{han_w3}
            \wordgroup[HAN_WORD]{1}{4}{4}{han_w4}

            \node[EN_WORD] (en_w1) [below=0pt of han_w1] {Shanghai};
            \node[EN_WORD] (en_w2) [below=0pt of han_w2] {plans to};
            \node[EN_WORD] (en_w3) [below=0pt of han_w3] {develop};
            \node[EN_WORD] (en_w4) [below=0pt of han_w4] {the financial sector};

        \end{dependency}
        \caption{A word-level dependency tree}
        \label{fig:word-tree}
    \end{subfigure}
    \hfill
    \begin{subfigure}[b]{1\columnwidth}
        \centering
        \begin{dependency}[]
            \begin{deptext}[column sep=2ex, font=\small]
            上 \& 海 \& 计 \& 划 \& 发 \& 展 \& 金 \& 融 \& 业  \\
            \end{deptext}
            \deproot[edge height=10ex]{4}{root}
            \depedge[intra word arc, hide label, edge height=2ex]{2}{1}{frag}
            \depedge[edge height=4.2ex]{4}{2}{nsubj}
            \depedge[intra word arc, hide label, edge height=2ex]{4}{3}{app}
            \depedge[edge height=4.2ex]{4}{5}{ccomp}
            \depedge[intra word arc, hide label, edge height=2ex]{5}{6}{coo}
            \depedge[intra word arc, hide label, edge height=3ex]{9}{7}{app}
            \depedge[intra word arc, hide label, edge height=2ex]{7}{8}{app}
            \depedge[edge height=4.2ex]{5}{9}{dobj}

            \node[EN_CHAR] (en_c1) [below=0pt of \wordref{1}{1}] {up};
            \node[EN_CHAR] (en_c2) [below=0pt of \wordref{1}{2}] {sea};
            \node[EN_CHAR] (en_c3) [below=0pt of \wordref{1}{3}] {plan};
            \node[EN_CHAR] (en_c4) [below=0pt of \wordref{1}{4}] {plot};
            \node[EN_CHAR] (en_c5) [below=0pt of \wordref{1}{5}] {grow};
            \node[EN_CHAR] (en_c6) [below=0pt of \wordref{1}{6}] {expand};
            \node[EN_CHAR] (en_c7) [below=0pt of \wordref{1}{7}] {gold};
            \node[EN_CHAR] (en_c8) [below=0pt of \wordref{1}{8}] {current};
            \node[EN_CHAR] (en_c9) [below=0pt of \wordref{1}{9}] {sector};
            
        \end{dependency}
        \caption{\citet{zhang-etal-2014-character}: annotated intra-word structure}
        \label{fig:annotated-structures}
    \end{subfigure} \\
    \begin{subfigure}[b]{1\columnwidth}
        \centering
        \begin{dependency}[]
            \begin{deptext}[column sep=2ex, font=\small]
            上 \& 海 \& 计 \& 划 \& 发 \& 展 \& 金 \& 融 \& 业  \\
            \end{deptext}
            \deproot[edge height=10ex]{4}{root}
            \depedge[intra word arc, hide label, edge height=2ex]{2}{1}{app}
            \depedge[edge height=4.2ex]{4}{2}{nsubj}
            \depedge[intra word arc, hide label, edge height=2ex]{4}{3}{app}
            \depedge[intra word arc, hide label, edge height=2ex]{6}{5}{app}
            \depedge[edge height=4.2ex]{4}{6}{ccomp}
            \depedge[intra word arc, hide label, edge height=2ex]{8}{7}{app}
            \depedge[intra word arc, hide label, edge height=2ex]{9}{8}{app}
            \depedge[edge height=4.2ex]{6}{9}{dobj}

            \node[EN_CHAR] (en_c1) [below=0pt of \wordref{1}{1}] {up};
            \node[EN_CHAR] (en_c2) [below=0pt of \wordref{1}{2}] {sea};
            \node[EN_CHAR] (en_c3) [below=0pt of \wordref{1}{3}] {plot};
            \node[EN_CHAR] (en_c4) [below=0pt of \wordref{1}{4}] {plan};
            \node[EN_CHAR] (en_c5) [below=0pt of \wordref{1}{5}] {launch};
            \node[EN_CHAR] (en_c6) [below=0pt of \wordref{1}{6}] {expand};
            \node[EN_CHAR] (en_c7) [below=0pt of \wordref{1}{7}] {gold};
            \node[EN_CHAR] (en_c8) [below=0pt of \wordref{1}{8}] {fuse};
            \node[EN_CHAR] (en_c9) [below=0pt of \wordref{1}{9}] {sector};
           
        \end{dependency}
        \caption{\citet{yan-etal-2020-graph}: pseudo leftward intra-word structure}
        \label{fig:leftward-structures}
    \end{subfigure} 
    \hfill
    \begin{subfigure}[b]{1\columnwidth}
        \centering
        \tikzset{root label style/.style = {
            anchor = mid,
            draw, solid,
            black,
            scale = .7,
            text height = 1.5ex, text depth = 0.25ex, %
            inner sep=.5ex, 
            outer sep = 0pt,
            rounded corners = 2pt, 
            text = black, 
            fill = white,
            draw=black},
            root edge style/.style = {->, >=stealth, black, solid, rounded corners = 2, line cap = round},
        }
        \begin{dependency}[]
            \begin{deptext}[column sep=2ex, font=\small]
            上 \& 海 \& 计 \& 划 \& 发 \& 展 \& 金 \& 融 \& 业  \\
            \end{deptext}
            \wordgroup[HAN_WORD]{1}{1}{2}{han_w1}
            \wordgroup[HAN_WORD]{1}{3}{4}{han_w2}
            \wordgroup[HAN_WORD]{1}{5}{6}{han_w3}
            \wordgroup[HAN_WORD]{1}{7}{9}{han_w4}

                        \node[EN_WORD] (en_w1) [below=0pt of han_w1] {};
            \node[EN_WORD] (en_w2) [below=0pt of han_w2] {};
            \node[EN_WORD] (en_w3) [below=0pt of han_w3] {};
            \node[EN_WORD] (en_w4) [below=0pt of han_w4] {};
        
            \node [WORD,fit={(han_w1) (en_w1)}] (w1) {};
            \node [WORD,fit={(han_w2) (en_w2)}] (w2) {};
            \node [WORD,fit={(han_w3) (en_w3)}] (w3) {};
            \node [WORD,fit={(han_w4) (en_w4)}] (w4) {};

             \node[EN_CHAR] (en_c1) [below=0pt of \wordref{1}{1}] {up};
            \node[EN_CHAR] (en_c2) [below=0pt of \wordref{1}{2}] {sea};
            \node[EN_CHAR] (en_c3) [below=0pt of \wordref{1}{3}] {plan};
            \node[EN_CHAR] (en_c4) [below=0pt of \wordref{1}{4}] {plot};
            \node[EN_CHAR] (en_c5) [below=0pt of \wordref{1}{5}] {grow};
            \node[EN_CHAR] (en_c6) [below=0pt of \wordref{1}{6}] {expand};
            \node[EN_CHAR] (en_c7) [below=0pt of \wordref{1}{7}] {gold};
            \node[EN_CHAR] (en_c8) [below=0pt of \wordref{1}{8}] {current};
            \node[EN_CHAR] (en_c9) [below=0pt of \wordref{1}{9}] {sector};
            
            \node[root label style] (root) [above=8ex of han_w2] {root};
            
            \node (iw1l) [above left of = w1, xshift=-0ex, yshift=2ex] {};
            \node (iw1r) [above right of = w1, xshift=0ex, yshift=-1.5ex] {};
            \draw [draw=none, fill=blue!20, thick, rounded corners=1mm] (iw1l) rectangle (iw1r);

            \node (iw2l) [above left of = w2, xshift=-0ex, yshift=2ex] {};
            \node (iw2r) [above right of = w2, xshift=0ex, yshift=-1.5ex] {};
            \draw [draw=none, fill=blue!20, thick, rounded corners=1mm] (iw2l) rectangle (iw2r);

            \node (iw3l) [above left of = w3, xshift=0ex, yshift=2ex] {};
            \node (iw3r) [above right of = w3, xshift=0ex, yshift=-1.5ex] {};
            \draw [draw=none, fill=blue!20, thick, rounded corners=1mm] (iw3l) rectangle (iw3r);

            \node (iw4l) [above left of = w4, xshift=-3ex, yshift=2ex] {};
            \node (iw4r) [above right of = w4, xshift=3ex, yshift=-1.5ex] {};
            \draw [draw=none, fill=blue!20, thick, rounded corners=1mm] (iw4l) rectangle (iw4r);

            \draw[root edge style, shorten >= 4ex] (root) -- ($(han_w2.north) + (0, -0ex)$);
            \groupedge[edge vertical padding=4ex]{han_w2}{han_w1}{nsubj}{3ex}
            \groupedge[edge vertical padding=4ex]{han_w2}{han_w3}{ccomp}{3ex}
            \groupedge[edge vertical padding=4ex]{han_w3}{han_w4}{dobj}{3ex}

            \depedge[<->, edge vertical padding=.8ex, edge start x offset=0.5ex, edge height=2ex, hide label, intra word arc]{2}{1}{app}
            \depedge[<->, edge vertical padding=.8ex, edge start x offset=-1ex, edge height=2ex, hide label, intra word arc]{3}{4}{app}
            \depedge[<->, edge vertical padding=.8ex, edge start x offset=-1ex, edge height=2ex, hide label, intra word arc]{5}{6}{app}
            \depedge[<->, edge vertical padding=.8ex, edge start x offset=-1ex, edge height=2ex, hide label, latent intra word arc]{7}{9}{app}
            \depedge[<->, edge vertical padding=.8ex, edge start x offset=-0.5ex, edge height=1.5ex, hide label, latent intra word arc]{7}{8}{app}
            \depedge[<->, edge vertical padding=.8ex, edge start x offset=0ex, edge end x offset=0.5ex, edge height=1.5ex, hide label, latent intra word arc]{9}{8}{app}

        \end{dependency}
        \caption{Ours: latent intra-word structure}\label{fig:latent-structures}
    \end{subfigure}\\
    \caption{
   A word-level dependency tree and corresponding character-level trees with three types of intra-word structure.
   Intra-word dependencies are represented by dashed arcs and their labels are omitted.
    }\label{fig:char-level-trees}
    
\end{figure*}

In the field of natural language processing, dependency parsing plays a crucial role in revealing the syntactic structure of sentences, 
thereby forming the foundation for numerous downstream applications such as machine translation \citep{shen-etal-2008-new,wu-etal-2017-sequence}, information extraction \citep{culotta-sorensen-2004-dependency,gamallo-etal-2012-dependency}, and sentiment analysis \citep{nakagawa-etal-2010-dependency,sun-etal-2019-aspect}. 

This task, although straightforward in space-delimited languages, encounters significant challenges in languages like Chinese, where explicit word boundaries are absent. 
Traditional Chinese parsing methods rely heavily on word-level treebanks, necessitating the segmentation of text into distinct words before parsing. 
This prerequisite not only adds an additional layer of complexity but also makes the parsing outcome vulnerable to inaccuracies in segmentation.

The need to address these issues has prompted a transition from word-level to character-level Chinese dependency parsing.
However, the lack of character-level Chinese treebanks presents a challenge.
As a workaround, researchers have endeavored to derive character-level dependency trees from word-level ones \citep{hatori-etal-2012-incremental,zhang-etal-2014-character,zhang-etal-2015-randomized,kurita-etal-2017-neural,li2018neural,yan-etal-2020-graph,wu2021deep}.

\citet{zhang-etal-2014-character} pioneered the integration of character- and word-level annotations. 
Figure~\ref{fig:char-level-trees} demonstrates a fully depicted character-level dependency tree (Figure~\ref{fig:annotated-structures}) by combining the word-level tree (Figure~\ref{fig:word-tree}) with annotated intra-word structures.
Nonetheless, the application of this method is constrained by the non-trivial task of deriving linguistically coherent intra-word structures.

Some researchers have opted for a simpler approach by defining pseudo intra-word structures \citep{hatori-etal-2012-incremental,yan-etal-2020-graph}.
As illustrated in Figure~\ref{fig:leftward-structures}, these structures utilize a left-wavy pattern, with the rightmost character acting as the root and other characters headed by their right-adjacent characters.
Although this method circumvents the labor-intensive annotation process, it may not accurately represent the syntactic roles of characters.

This paper proposes a new approach to character-level Chinese dependency parsing via modeling latent intra-word structure. 
As illustrated in Figure~\ref{fig:latent-structures}, our approach allows for the implicit representation of all potential internal structures within words.
For example, for the word ``发展 (develop)'', both ``发 (grow)$\rightarrow$展 (expand)'' and ``发 (grow)$\leftarrow$展 (expand)'' are acceptable structures. 
In this way, \emph{each word-level dependency tree is interpreted as a forest of character-level trees.}

Central to our approach is a constrained Eisner algorithm \citep{eisner-1996-three}, crafted to maintain the compatibility of character-level trees it generates. 
This algorithm enforces two critical constraints: the \emph{single-root subtree} constraint and the \emph{root-as-head} constraint, which together guarantee that each word corresponds to a single-root subtree and that inter-word dependencies link to root characters of subtrees. 
Furthermore, we introduce a coarse-to-fine parsing strategy to refine the parsing process.
Our primary contributions include:
\begin{itemize}[leftmargin=*]
        \item This work explores modeling latent intra-word structure for character-level Chinese dependency parsing.
        \item  By implementing a novel, linguistically informed algorithm, the compatibility of character-level trees with their word-level counterparts is ensured.
        \item We devise a coarse-to-fine parsing strategy that improves parsing accuracy and generates more linguistically plausible intra-word structures.
        \item Experiment results on Chinese treebanks demonstrate that our approach outperforms both the pipeline model and previous joint models. Additionally, we provide insightful analyses of the predicted intra-word structures.

\end{itemize}
We will release our code at \url{https://github.com/ironsword666/CharDepParsing}.

\section{Parsing with Latent Structure}\label{section:span-based-constraints}

\subsection{Word-level Tree to Char-level Forest}\label{sec:word-to-char}

\paragraph{Latent Structure.} To transform word-level trees into character-level trees, previous studies typically defined fixed internal structures for each word, either annotated by human experts \citep{zhang-etal-2014-character} or generated through rules \citep{yan-etal-2020-graph}.
Our approach does not explicitly define intra-word structures.
Instead, it allows for the representation of all possible internal structures within each word.
This method acknowledges the multifaceted nature of language, where a single word may have multiple structures, especially for words with multiple parts of speech and coordinate characters \citep{gong-etal-2021-depth}. 
The implicit representation of intra-word structures empowers the model to identify the most plausible structure based on context.

\paragraph{Conversion.} The latent nature of the intra-word structures facilitates a flexible construction of character-level dependencies, which are categorized into intra-word and inter-word for clarity. 
Within a given word, any two characters can form an intra-word dependency.
Conversely, given a head-modifier pair, an inter-word dependency can originate from any characters in the head word to any characters in the modifier word.
In this way, a word-level dependency tree can be interpreted as a forest comprising various potential character-level trees,
as illustrated by the specific examples in Figures~\ref{fig:annotated-structures} and~\ref{fig:leftward-structures} for the forest in Figure~\ref{fig:latent-structures}.

\subsection{Compatibility: Two Constraints}\label{sec:two-constaints}

The aforementioned conversion process is structurally sound, indicating there are no conflicts between dependencies in the converted character-level trees and dependencies in the original word-level trees.
However, ensuring the character-level trees faithfully represent both the internal structure of words and the syntactic relationships between them requires addressing compatibility issues. These issues, while not explicitly defined, adhere to certain linguistic principles.
To this end, we introduce two constraints:
\begin{figure}[tb]
    \centering
    \footnotesize
    \begin{subfigure}[t]{0.45\columnwidth}
        \centering
        \begin{tikzpicture}[>=stealth, scale=0.7]
            \node (c1) at (-2, 0) [above] {发};
            \node (c2) at (-1, 0) [above] {展};
            \node[fill=red!20] (c3) at (0, 0) [above] {金};
            \node (c4) at (1, 0) [above] {融};
            \node[fill=red!20] (c5) at (2, 0) [above] {业};
            \coordinate (c23) at ($(c2)!0.5!(c3)$);
            \draw[solid] (c1.south west) -- (c5.south east);
            \draw[solid] (c1.south west) -- (c1.south west |- c23);
            \draw[solid] (c23 |- c3.south) -- (c23);
            \draw[solid] (c5.south east) -- (c5.south east |- c23);
            \node (c451) at ($(c4)!0.5!(c5) + (-1ex, 0)$) [] {};
            \node (c452) at ($(c4)!0.5!(c5)$) [] {};
            \node (C1_end) at (c451 |- c4.north) [] {};
            \node [coordinate] (C1) [above=10ex of c1.north] {};
            \node [coordinate] (C2) [above=3ex of c5.north] {};
            \node [coordinate] (C31) [above=4.5ex of c3.north, xshift=-0.2ex] {};
            \node [coordinate] (C32) [above=4.5ex of c3.north, xshift=0.2ex] {};
            \draw[thick, ] (C1) -- (c1.north) -- (C31 |- c3.north) -- (C31) -- cycle;
            \draw[thick, densely dashed] (C32) -- (C32 |- c3.north) -- (c451 |- c4.north) -- cycle;
            \draw[thick, densely dashed] (C2) -- (C2.north |- c1.north) -- (c452 |- c4.north) -- cycle;
            \draw[->, thick, red] (c1.north) to[bend left=60] ($(c5.north) + (-0.2ex, 0)$);
            \draw[->, thick, densely dashed] (c1.north) to[bend left=60] (c2.north);
            \draw[->, thick, red] (c1.north) to[bend left=60] ($(c3.north) + (-0.5ex, 0)$);
            \draw[->, thick, densely dashed] ($(c3.north) + (0.2ex, 0)$) to[bend left=60] (c4.north);

            \node[text height=1ex, text depth=.5ex, anchor=north, font=\tiny] at (c1.south) {grow};
            \node[text height=1ex, text depth=.5ex, anchor=north, font=\tiny] at (c2.south) {expand};  
            \node[text height=1ex, text depth=.5ex, anchor=north, font=\tiny] at (c3.south) {gold};
            \node[text height=1ex, text depth=.5ex, anchor=north, font=\tiny] at (c4.south) {current};
            \node[text height=1ex, text depth=.5ex, anchor=north, font=\tiny] at (c5.south) {sector};

        \end{tikzpicture}
        \caption{Two root characters: ``金(gold)'' and ``业(sector)''}\label{fig:multi-root}
    \end{subfigure}
    \hfill
    \begin{subfigure}[t]{0.45\columnwidth}
        \centering
        \begin{tikzpicture}[>=stealth, scale=0.7]
            \node (c1) at (-2, 0) [above] {发};
            \node[fill=red!20] (c2) at (-1, 0) [above] {展};
            \node (c3) at (0, 0) [above] {金};
            \node (c4) at (1, 0) [above] {融};
            \node (c5) at (2, 0) [above] {业};
            \coordinate (c23) at ($(c2)!0.5!(c3)$);
            \draw[] (c1.south west) -- (c5.south east);
            \draw[] (c1.south west) -- (c1.south west |- c23);
            \draw[] (c23 |- c3.south) -- (c23);
            \draw[] (c5.south east) -- (c5.south east |- c23);
            \node[coordinate] (C1) [above=10ex of c1.north] {};
            \node[coordinate] (C2) [above=7.7ex of c2.north] {};
            \node[coordinate] (C21) [above=7.7ex of c2.north, xshift=-0.2ex] {};
            \node[coordinate] (C22) [above=7.7ex of c2.north, xshift=0.2ex] {};
            \draw[thick, densely dashed] (C1) -- (C21) --  (C21 |- c2.north) -- (c1.north) -- cycle;
            \draw[thick] (C22) -- (C22 |- c2.north) -- (c5.north east) -- cycle;
            \draw[->, thick, densely dashed, out=60, in=120] ($(c1.north) + (0, 0ex)$) to[bend left=60] ($(c2.north) + (-0.5ex, 0.1ex)$);
            \draw[->, out=45, in=135, thick, red] ($(c2.north) + (0.5ex, 0ex)$) to ($(c5.north) + (-0.5ex, 0ex)$);
            \draw[->, thick, densely dashed, out=135, in=45] ($(c5.north) + (-1.5ex, 0ex)$) to ($(c3.north) + (-0.5ex, 0ex)$);
            \draw[->, thick, densely dashed, out=60, in=120] ($(c3.north) + (0.5ex, 0ex)$) to ($(c4.north) + (0, 0ex)$);

            \node[text height=1ex, text depth=.5ex, anchor=north, font=\tiny] at (c1.south) {grow};
            \node[text height=1ex, text depth=.5ex, anchor=north, font=\tiny] at (c2.south) {expand};  
            \node[text height=1ex, text depth=.5ex, anchor=north, font=\tiny] at (c3.south) {gold};
            \node[text height=1ex, text depth=.5ex, anchor=north, font=\tiny] at (c4.south) {current};
            \node[text height=1ex, text depth=.5ex, anchor=north, font=\tiny] at (c5.south) {sector};

        \end{tikzpicture}
        \caption{Non-root character ``展(expand)'' as head}\label{fig:multi-out}
    \end{subfigure}
    \\
    \begin{subfigure}[t]{\columnwidth}
        \centering
        \begin{tikzpicture}[>=stealth, scale=1]
            \node (c1) at (-2, 0) [above] {发};
            \node (c2) at (-1, 0) [above] {展};
            \node (c3) at (0, 0) [above] {金};
            \node (c4) at (1, 0) [above] {融};
            \node (c5) at (2, 0) [above] {业};
            \coordinate (c23) at ($(c2)!0.5!(c3)$);
            \draw[solid] (c1.south west) -- (c5.south east);
            \draw[solid] (c1.south west) -- (c1.south west |- c23);
            \draw[solid] (c23 |- c3.south) -- (c23);
            \draw[solid] (c5.south east) -- (c5.south east |- c23);
         
            \draw[->, thick, black] (c1.north) to[bend left=60] ($(c5.north) + (0.5ex, 0)$);
            \draw[->, thick, densely dashed] (c1.north) to[bend left=60] (c2.north);
            \draw[->, thick, densely dashed, black]  ($(c5.north) + (-0.5ex, 0)$) to[bend right=60] ($(c3.north) + (-0.7ex, 0)$);
            \draw[->, thick, red] ($(c3.north) + (0.2ex, 0)$) to[bend left=60] (c4.north);

            \node[text height=1ex, text depth=.5ex, anchor=north, font=\tiny] at (c1.south) {grow};
            \node[text height=1ex, text depth=.5ex, anchor=north, font=\tiny] at (c2.south) {expand};  
            \node[text height=1ex, text depth=.5ex, anchor=north, font=\tiny] at (c3.south) {gold};
            \node[text height=1ex, text depth=.5ex, anchor=north, font=\tiny] at (c4.south) {current};
            \node[text height=1ex, text depth=.5ex, anchor=north, font=\tiny] at (c5.south) {sector};

        \end{tikzpicture}
        \caption{An inter-word arc underlies an intra-word arc}\label{fig:inter-under-intra}
    \end{subfigure}
    \caption{
        Examples containing illegal arcs.
        Incorrect characters and arcs are highlighted in red.
        Triangles represent complete spans, while trapezoids represent incomplete spans. 
        Dashed or solid lines are used to indicate intra-word or inter-word.
        }\label{fig:illegal-tree}
\end{figure}
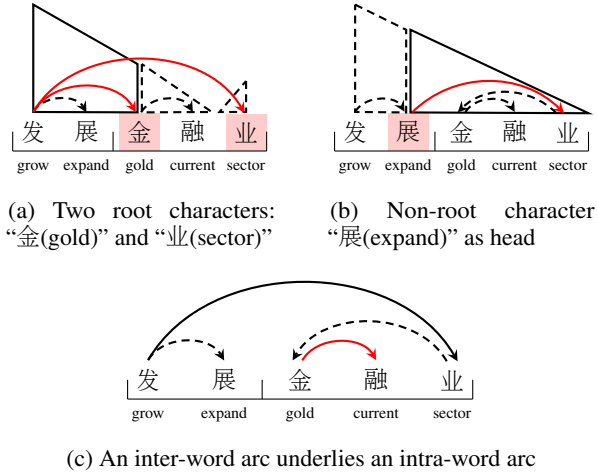

\textbf{(1) The single-root subtree constraint.} This constraint upholds the linguistic principle that each word corresponds to a single-root subtree within the character-level trees. 
It implies several aspects:
\begin{enumerate*}[label=(\roman*)]
    \item characters in the word form a subtree;
    \item there is a single, most important character representing the word, selected as the root of the subtree;
    \item all other characters are descendants of this root character;
    \item given the single-headed nature of dependency trees, the root character—and only the root character—can modify a character from another word, resulting in an inter-word dependency. 
\end{enumerate*}
An illustration showing a word erroneously assigned two root characters is provided in Figure~\ref{fig:multi-root}.

\textbf{(2) The root-as-head constraint.} 
While the single-root subtree constraint guarantees that only the root character can act as the modifier in an inter-word dependency,
it is possible that a root character of one word modifies a non-root character in another word, as shown in Figure~\ref{fig:multi-out}.
To accurately reflect the relation between intra-word structures, we require that only the root character of a word can serve as the head in an inter-word dependency.

The two constraints collectively assert that 
\emph{a root character not only represents the central syntactic role of 
the word but also exclusively participates in forming inter-word dependencies}.

\subsection{The Constrained Eisner Algorithm}\label{sec:constrained-decoding}
\paragraph{Outline.} 
This work integrates the proposed constraints into both the Eisner algorithm \citep{eisner-1996-three} and the Inside algorithm \citep{eisner-2016-inside}. During training, we employ the constrained Inside algorithm on word-level trees to compute the training loss. During inference, the vanilla Eisner algorithm is applied to character sequences to derive optimal character-level trees, while the constrained Eisner algorithm is used on word sequences for analytical purposes.
In the following part, we use the Eisner algorithm as an example to demonstrate the implementation of these constraints.
\paragraph{Eisner algorithm.}
The Eisner algorithm, a classic dynamic programming approach, iteratively combines smaller spans into larger ones to construct a complete dependency tree. It distinguishes between two types of spans: complete spans and incomplete spans. Complete spans consist of a head word and all its descendants located on one side, while incomplete spans include a head-modifier dependency and the region between the head and the modifier. 

Given all scores of character-level dependencies, obtaining an optimal character-level tree using the vanilla Eisner algorithm is straightforward. However, deriving an optimal character-level tree that is \emph{compatible} with a given word sequence is more complex. To address this, we propose a constrained Eisner algorithm, detailed in Algorithm~\ref{alg:variant-inside}.\footnote{Similarly, the constrained Inside algorithm can be implemented by replacing max-product with sum-product.}

\begin{algorithm}[tb]
    \caption{Constrained Eisner Algorithm.}\label{alg:variant-inside}
    {\small
        \begin{algorithmic}[1]
        \State \textbf{Input}: arc scores $s(i,j)$
        \LComment{arc scores conflicting with gold-standard segmentation are masked to $-\infty$}
        \State \textbf{Define}: $I,C \in \mathbf{R}^{n \times n}$ 
        \State \textbf{Initialize}: $C_{i \rightarrow i} = 0, 1 \leq i \leq n$
        \For {$w = 1, \ldots, \ n$}
            \For {$i = 1, \ldots, \ n - w$}
                \State {$j = i + w$}
                \State $\begin{aligned}
                    I_{i \rightarrow j} = \max_{i \le k < j}(s(i,j) + C_{i \rightarrow k} + C_{k+1 \leftarrow j})
                \end{aligned}$ 
                \State $\begin{aligned}
                    I_{i \leftarrow j} = \max_{i \le k < j}(s(j,i) + C_{i \rightarrow k} + C_{k+1 \leftarrow j})
                \end{aligned}$ 
                \State $\begin{aligned}
                    C_{i \rightarrow j} = \max_{i < k \le j}(I_{i \rightarrow k} + C_{k \rightarrow j})
                \end{aligned}$
                \LComment{j belongs to the right boundaries of the words}
                \LComment{if (i,k) inside a word, (k,j) also inside this word}
                \State $\begin{aligned}
                    C_{i \leftarrow j} = \max_{i \le k < j}(C_{i \leftarrow k} + I_{k \leftarrow j})
                \end{aligned}$
                \LComment{i belongs to the left boundaries of the words}
                \LComment{if (k,j) inside a word, (i,k) also inside this word}
                
            \EndFor
        \EndFor
        \State \Return  $C_{1 \rightarrow n}$
        \end{algorithmic}
    }
\end{algorithm}

\paragraph{Constraint enforcement.}
To clarify the implementation of two constraints, we first differentiate spans into two types: intra-word and inter-word.
Intra-word spans consist solely of intra-word dependencies, spanning either part or the entirety of a word.
Inter-word spans contain at least one inter-word dependency, spanning multiple words.  
Please refer to the examples in Figure~\ref{fig:illegal-tree}.

For the single-root subtree constraint, we observe that cases of multi-roots arise from inter-word complete spans including residual characters from a word (see Figure~\ref{fig:multi-root} for an example). 
Inspired by recent work \citep{zhang-etal-2021-adapting,zhang-etal-2022-semantic}, we stipulate that inter-word complete spans must terminate at word boundaries.

For the root-as-head constraint, based on our observations, instances where non-characters become the heads of inter-word dependencies arise when combining an intra-word incomplete span with an inter-word complete span.
An example is provided in Figure~\ref{fig:multi-out}.
Therefore, we prohibit all such combination operations.
To the best of our knowledge, \emph{we are the first to address the root-as-head constraint in graph-based dependency parsing.}

The implementation of two constraint rules is straightforward by using auxiliary mask tensors.
The additional time complexity is $O(n^3)$ but becomes negligible when accelerated by GPUs.

\subsection{A Coarse-to-Fine Parsing Strategy}\label{subsection:coarse2fine-decoding}
In the absence of the word-level trees, determining the intra-word and inter-word roles for a dependency in the character-level trees is not straightforward.
Since the Eisner algorithm conflates two distinct roles, identifying these roles is only possible after the arc labeling step (described in Section~\ref{sec:model}).
This can lead to instances where an intra-word dependency arc overlies an inter-word dependency arc (see Figure~\ref{fig:inter-under-intra}).
These illegal arcs hinder the recovery from character-level trees to word-level trees (see Appendix~\ref{appendix:tree-recovery} for details).

To ensure the validity of output trees, we propose a coarse-to-fine parsing strategy, explicitly assigning each arc two scores for intra-word and inter-word roles.
The core idea is to first construct intra-word spans and then inter-word spans, thus ensuring that intra-word dependency arcs underlie the inter-word dependency arcs.
The deduction rules are depicted in Figure~\ref{fig:c2f-deduction}.
We refer interested readers to Algorithm~\ref{alg:c2f-eisner} in the appendix for details.

\begin{figure}[tb!]
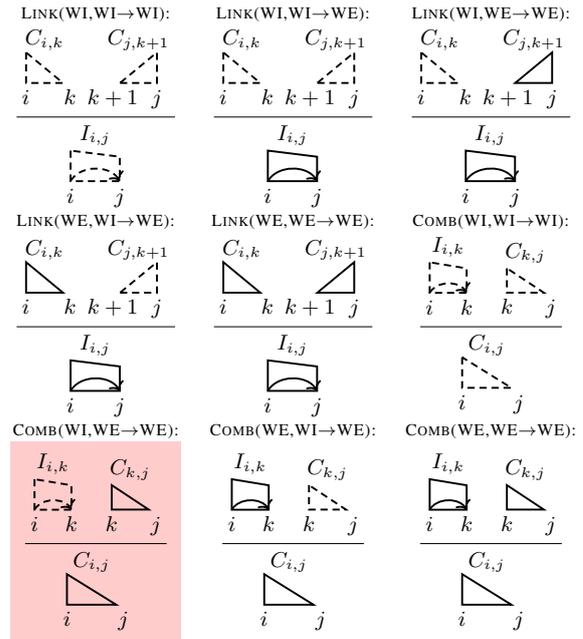

    \definecolor{seagreen}{HTML}{6bc1a2}
    \newcommand{\seagreen}[1]{\textcolor{seagreen}{#1}}
    \renewcommand{\arraystretch}{1}
    \centering
    \scalebox{0.9}{
        \scriptsize
        \begin{tabular}{ccc}
            \textsc{Link(WI,WI$\to$WI):} &  \textsc{Link(WI,WI$\to$WE):} & \textsc{Link(WI,WE$\to$WE):} \\
            $\inferrule{
                    \tikz[baseline=0pt]{\righttriangledash[0.6][0.5]{i}{k}{C_{i,k}}}
                    \tikz[baseline=0pt]{\lefttriangledash[0.6][0.5]{k+1}{j}{C_{j,k+1}}}
                } {
                    \tikz[baseline=0pt]{\trapezoidarcdash[0.8][0.5][0.1]{i}{j}{I_{i,j}}}
                }$  &
            $\inferrule{
                    \tikz[baseline=0pt]{\righttriangledash[0.6][0.5]{i}{k}{C_{i,k}}}
                    \tikz[baseline=0pt]{\lefttriangledash[0.6][0.5]{k+1}{j}{C_{j,k+1}}}
                } {
                    \tikz[baseline=0pt]{\trapezoidarc[0.8][0.5][0.1]{i}{j}{I_{i,j}}}
                }$  &
            $\inferrule{
                    \tikz[baseline=0pt]{\righttriangledash[0.6][0.5]{i}{k}{C_{i,k}}}
                    \tikz[baseline=0pt]{\lefttriangle[0.6][0.5]{k+1}{j}{C_{j,k+1}}}
                } {
                    \tikz[baseline=0pt]{\trapezoidarc[0.8][0.5][0.1]{i}{j}{I_{i,j}}}
                }$ \\
            \textsc{Link(WE,WI$\to$WE):} &  \textsc{Link(WE,WE$\to$WE):} & \textsc{Comb(WI,WI$\to$WI):} \\
            $\inferrule{
                    \tikz[baseline=0pt]{\righttriangle[0.6][0.5]{i}{k}{C_{i,k}}}
                    \tikz[baseline=0pt]{\lefttriangledash[0.6][0.5]{k+1}{j}{C_{j,k+1}}}
                } {
                    \tikz[baseline=0pt]{\trapezoidarc[0.8][0.5][0.1]{i}{j}{I_{i,j}}}
                }$  &
            $\inferrule{
                    \tikz[baseline=0pt]{\righttriangle[0.6][0.5]{i}{k}{C_{i,k}}}
                    \tikz[baseline=0pt]{\lefttriangle[0.6][0.5]{k+1}{j}{C_{j,k+1}}}
                } {
                    \tikz[baseline=0pt]{\trapezoidarc[0.8][0.5][0.1]{i}{j}{I_{i,j}}}
                }$  &
            $\inferrule{
                    \tikz[baseline=0pt]{\trapezoidarcdash[0.6][0.5][0.1]{i}{k}{I_{i,k}}} \quad
                    \tikz[baseline=2.5pt]{\righttriangledash[0.6][0.4]{k}{j}{C_{k,j}}}
                } {
                    \tikz[baseline=0pt]{\righttriangledash[0.8][0.5]{i}{j}{C_{i,j}}}
                }$ \\
            \textsc{Comb(WI,WE$\to$WE):} &  \textsc{Comb(WE,WI$\to$WE):} & \textsc{Comb(WE,WE$\to$WE):} \\
            \colorbox{red!20}{
                $\inferrule{
                        \tikz[baseline=0pt]{\trapezoidarcdash[0.6][0.5][0.1]{i}{k}{I_{i,k}}} \quad
                        \tikz[baseline=2.5pt]{\righttriangle[0.6][0.4]{k}{j}{C_{k,j}}}
                    } {
                        \tikz[baseline=0pt]{\righttriangle[0.8][0.5]{i}{j}{C_{i,j}}}
                    }$ 
                } &
            $\inferrule{
                    \tikz[baseline=0pt]{\trapezoidarc[0.6][0.5][0.1]{i}{k}{I_{i,k}}} \quad
                    \tikz[baseline=2.5pt]{\righttriangledash[0.6][0.4]{k}{j}{C_{k,j}}}
                } {
                    \tikz[baseline=0pt]{\righttriangle[0.8][0.5]{i}{j}{C_{i,j}}}
                }$  &
            $\inferrule{
                    \tikz[baseline=0pt]{\trapezoidarc[0.6][0.5][0.1]{i}{k}{I_{i,k}}} \quad
                    \tikz[baseline=2.5pt]{\righttriangle[0.6][0.4]{k}{j}{C_{k,j}}}
                } {
                    \tikz[baseline=0pt]{\righttriangle[0.8][0.5]{i}{j}{C_{i,j}}}
                }$ \\
        \end{tabular}
    }
    \caption{
    Deduction rules for coarse-to-fine parsing.
    Dashed or solid lines are used to indicate intra-word spans (WI) or inter-word spans (WE).
    The highlighted rule can be ignored to satisfy the \emph{root-as-head} constraint.
    We present only R-rules, omitting the symmetric L-rules and initial conditions for brevity.\label{fig:c2f-deduction}}
\end{figure}

\section{Model}\label{sec:model}

\paragraph{Notations.} Given a sentence $\boldsymbol{x} = c_0c_1\ldots c_n $, where $c_i$ represents the $i$th character of $\boldsymbol{x}$ and $c_0$ denotes an artificial $\textsc{Root}$ token, a labeled dependency tree for $\boldsymbol{x}$ is denoted as $\boldsymbol{t}$. 
We view $\boldsymbol{t}$ as a set of labeled dependency arcs, using $(i,j,l) \in \boldsymbol{t}$ to indicate an arc from character $c_i$ to $c_j$ with a label $l \in \mathcal{L}$, where $\mathcal{L}$ is the set of dependency labels.\footnote{An additional $\textsc{Intra}$ label is used to indicate the intra-word dependency arcs.} 
Additionally, an unlabeled dependency tree is denoted as $\boldsymbol{y}$ and an unlabeled dependency arc is denoted as $(i,j)$. 

\subsection{Parsing Modeling} 
Adhering to \citet{DBLP:conf/iclr/DozatM17}, we employ a two-stage parsing framework that first predicts unlabeled trees and then labels the arcs in these trees. 
The score of an unlabeled dependency tree is the cumulative sum of its unlabeled arc scores:
\begin{equation} \label{equation:unlabeled tree score}
    s(\boldsymbol{x},\boldsymbol{y}) = \sum\limits_{(i,j) \in \boldsymbol{y}}{s(i,j)}
\end{equation}

The conditional probability of a unlabeled tree $\boldsymbol{y}$ is defined as:
\begin{equation} \label{equation:tree-prob}
    p(\boldsymbol{y}|\boldsymbol{x}) =  \frac
    {e^{s(\boldsymbol{x},\boldsymbol{y})}}
    {\boldsymbol{Z(x)}\equiv\sum\limits_{\boldsymbol{y'} \in \mathcal{Y}(x)}{e^{s(\boldsymbol{x},\boldsymbol{y}')}}}
\end{equation}
where $\boldsymbol{Z(x)}$ is known as the partition term, and $\mathcal{Y}(x)$ denotes the set of all possible (projective) trees for $\boldsymbol{x}$.

\paragraph{Forest probability.} The forest, denoted as $\mathcal{F}$, comprises dependency trees that meet compatibility constraints. 
The probability of $\mathcal{F}$ is the aggregate probability of each tree $\boldsymbol{y}$ within $\mathcal{F}$.
\begin{equation} \label{equation:forest-prob}
\begin{aligned}
    p(\mathcal{F}|\boldsymbol{x}) 
    &= \sum\limits_{\boldsymbol{y} \in \mathcal{F}} p(\boldsymbol{y}|\boldsymbol{x}) \\
    &= \frac
    {\boldsymbol{Z(x,\mathcal{F})}\equiv\sum\limits_{\boldsymbol{y} \in \mathcal{F}} e^{s(\boldsymbol{x},\boldsymbol{y})}}
    {\boldsymbol{Z(x)}} 
\end{aligned}
\end{equation}
where $\boldsymbol{Z(x,\mathcal{F})}$ can be computed via a constrained Inside algorithm, by substituting the max-product in Algorithm~\ref{alg:variant-inside} with the sum-product.

\subsection{Training}
During training, the loss for a sentence $\boldsymbol{x}$ is composed of two parts: (unlabeled) tree loss and label loss.
\begin{equation}
    \mathbf{\emph{L}}(\boldsymbol{x}) = \mathbf{\emph{L}}^{\mathrm{tree}}(\boldsymbol{x}) + \mathbf{\emph{L}}^{\mathrm{label}}(\boldsymbol{x})
\end{equation}

\paragraph{Tree loss.} Given a sentence $\boldsymbol{x}$,
the tree loss is naturally defined as the negative log-probability of the forest $\mathcal{F}$:
\begin{equation} \label{tree-loss}
    \mathbf{\emph{L}}^{\mathrm{tree}}(\boldsymbol{x}) 
    = -\log p(\mathcal{F}|\boldsymbol{x}) 
\end{equation}

\paragraph{Label loss.} The probability of assigning label $l$ to an unlabeled arc $(i,j)$ is defined as:
\begin{equation}\label{equation:label-prob}
p(l | i,j) =   \frac{e^{s(i,j,l)}}{
        \sum_{l' \in \mathcal{L}}{e^{s(i,j,l')}}}
\end{equation}

The label loss is the sum of negative log probabilities of correctly labeling each arc in the forest $\mathcal{F}$.\footnote{Refer to Appendix~\ref{appendix:loss-function} for the enumeration of these arcs.}
\begin{equation}\label{equation:multiplication-label-loss}
    \mathbf{\emph{L}}^{\mathrm{label}}(\boldsymbol{x}) = 
    \sum_{\boldsymbol{y} \in \mathcal{F}} \sum_{(i,j) \in \boldsymbol{y}} -\log p(l | i,j)
\end{equation}

\subsection{Inference}
To parse a sentence $\boldsymbol{x}$, the model first selects the highest-scoring unlabeled tree $\hat{\boldsymbol{y}}$ via (vanilla) Eisner algorithm.
\begin{equation} \label{equation:max-score-tree}
    \boldsymbol{\hat{y}} = \mathop{\arg\max}_{\boldsymbol{y} \in \mathcal{Y}(\boldsymbol{x})}{s(\boldsymbol{x},\boldsymbol{y})}
\end{equation}
Subsequently, the optimal label for each arc $(i,j) \in \boldsymbol{\hat{y}}$ is determined.
\begin{equation} \label{equation:max-fine-label}
    \hat{l} = \mathop{\arg\max}_{l \in \mathcal{L}}{s(i,j,l)}
\end{equation}

\subsection{Network Architecture}
\paragraph{Encoding.} 
The sentence $\boldsymbol{x}$ is directly input into the pre-trained BERT model, and the output from the last layer is used as the representation of characters.
\begin{equation}
    \ldots,\mathbf{h}_i,\dots=\mathbf{BERT}(\dots,c_i,\dots)
\end{equation}

\paragraph{Scoring.}
To score dependency arcs, we utilize the biaffine attention mechanism as outlined by \citet{DBLP:conf/iclr/DozatM17}.
In the coarse-to-fine parsing, intra- and inter-word arcs are scored separately through distinct biaffine attentions. More details are provided in Appendix~\ref{appendix:c2f-scoring}.

\section{Experiments}

\begin{table*}[tb]
    \centering
    \begin{tabular}{lccccccccc}
    \toprule
    \multirow{3}{*}[0.75ex]{Model}  & \multicolumn{3}{c}{CTB5} & \multicolumn{3}{c}{CTB6} & \multicolumn{3}{c}{CTB7} \\\cmidrule(lr){2-4}\cmidrule(lr){5-7}\cmidrule(lr){8-10}
    & F1$_{seg}$ & UF$_{dep}$ & LF$_{dep}$ & F1$_{seg}$ & UF$_{dep}$ & LF$_{dep}$ & F1$_{seg}$ & UF$_{dep}$ & LF$_{dep}$\\
    \midrule
    \multicolumn{10}{c}{\emph{w/ head-finding rules of SD} } \\[2pt]
    \citet{yan-etal-2020-graph}  & 98.46 & 89.59 & 85.94 & --    & --  & -- & 97.06 & 85.06 & 80.71\\
    \xmidrule[10]
    Pipeline & 98.72 & 90.93 & 88.39 & 97.23 & 87.09 & 83.86 & 97.16 & 85.77 & 82.00 \\
    Leftward & 98.76 & 90.91 & 88.37 & 97.30 & 87.21 & 84.04 & \textbf{97.22} & 85.85 & 82.17 \\
    Latent (\emph{Ours})  & 98.76 & \textbf{91.06} & \textbf{88.49} & 97.28 & 87.22 & 84.03 & 97.17 & 85.74 & 82.04 \\
    Latent-c2f (\emph{Ours}) & \textbf{98.79} & 90.95 & 88.34 & \textbf{97.33} & \textbf{87.30} & \textbf{84.12} & \textbf{97.22} & \textbf{85.90} & \textbf{82.23} \\
    \midrule
    \multicolumn{10}{c}{\emph{w/ head-finding rules of Z\&C} } \\[2pt]
    \citet{hatori-etal-2012-incremental}$^{\dagger}$     & 97.75 & 81.56  & -- &  95.45  & 74.88  & -- & 95.42 & 73.58 & -- \\
    \citet{zhang-etal-2014-character}$^{\dagger}$        & 97.67    & 81.63  & -- & 95.63    & 76.75  & -- & 95.53 & 75.63  & -- \\
    \citet{zhang-etal-2015-randomized}$^{\dagger}$        &  98.04  & 82.01  & --& --    & --  & -- & --    & --  & --\\
    \citet{kurita-etal-2017-neural}$^{\dagger}$        & 98.37    & 81.42  & --& --    & --  & -- & 95.86    & 74.04  & --\\
    \citet{wu2021deep}$^{\ddagger}$ & 98.57 & 91.79 & 90.38 & 97.32 & 88.44 & 86.49 & 97.25   & 86.93 & 84.68 \\
    \xmidrule[10]
    Pipeline & 98.72 & 92.00 & 91.04 & 97.23 & 87.71 & 86.77 & 97.16 & 86.39 & 85.19 \\
    Leftward & 98.67 & 91.98 & 91.03 & \textbf{97.39} & 88.02 & 87.06 & \textbf{97.26} & 86.48 & 85.30 \\
    Latent (\emph{Ours}) & 98.74 & \textbf{92.16} & \textbf{91.25} & 97.37 & 87.99 & 87.06 & 97.22 & 86.50 & 85.33 \\
    Latent-c2f (\emph{Ours}) & \textbf{98.77} & 92.03 & 91.08 & 97.37 & \textbf{88.10} & \textbf{87.14} & \textbf{97.26} & \textbf{86.55} & \textbf{85.37} \\
    \bottomrule
    \end{tabular}
    
    \caption{Results on CTB5, CTB6, and CTB7 test sets. 
            The best results are in bold.
            $\dagger$ indicates using additional POS tag information. 
            $\ddagger$:  \citet{wu2021deep} consider a dependency arc correct even if the head word is wrongly segmented; thus, the reported results are not directly comparable to ours.}\label{table:joint-parsing}
\end{table*}

\paragraph{Data.} 
We conduct experiments on three versions of the Penn Chinese Treebank (CTB): CTB5, CTB6, and CTB7.\footnote{\href{https://catalog.ldc.upenn.edu/LDC2010T07}{https://catalog.ldc.upenn.edu/LDC2010T07}}
The split of train, development, and test sets follows established practices \citep{zhang-clark-2010-fast,yang-xue-2012-chinese,wang-etal-2011-improving}. 
Table~\ref{table:data-statistic} in the appendix provides detailed statistics.
The conversion from phrase structures to dependency structures is performed using two methods:
\begin{enumerate*}[label=(\arabic*)]
    \item the Stanford parser v3.3.0\footnote{\href{https://nlp.stanford.edu/software/lex-parser.shtml}{https://nlp.stanford.edu/software/lex-parser.shtml}} with Stanford Dependencies (SD) \citep{de-marneffe-etal-2006-generating};
    \item the Penn2Malt tool\footnote{\href{https://cl.lingfil.uu.se/\textasciitilde nivre/research/Penn2Malt.html}{https://cl.lingfil.uu.se/\textasciitilde nivre/research/Penn2Malt.html}} with the head-finding rules as described by \citet{zhang-clark-2008-tale}, henceforth referred to as Z\&C.
\end{enumerate*}
Only projective trees are retained during training.
An intra-word structure dataset annotated by \citet{gong-etal-2021-depth} on CTB5 is utilized for experiments and analysis.\footnote{\href{https://github.com/SUDA-LA/wist}{https://github.com/SUDA-LA/wist}} 

\paragraph{Evaluation metrics.}
For Chinese word segmentation (CWS), we employ standard F1 measures (F1$_{seg}$).
For dependency parsing, evaluation is conducted at the word level, using word-level F1 scores (UF$_{dep}$ and LF$_{dep}$) as the evaluation metrics \citep{yan-etal-2020-graph}.
A dependency arc is considered correct only if the head-modifier word pair is correctly segmented.
Punctuation is excluded during the evaluation of dependency parsing.

\paragraph{Baseline and proposed models.}
The evaluation includes the following models:
\begin{itemize}[leftmargin=*]
    \item \textbf{TreeCRF}: A word-level biaffine parsing model with a CRF loss, detailed in \citet{zhang-etal-2020-efficient}.
    \item \textbf{Pipeline}: This framework first performs CWS by assigning `BMES' tags to characters and then feeds the segmented results into \textbf{TreeCRF}.
    \item \textbf{Leftward}:  A model uses pseudo leftward intra-word structures as described by \citet{yan-etal-2020-graph}.
    \item \textbf{Latent}:  The proposed model uses latent intra-word structures. The constrained Eisner algorithm is used to ensure compatibility.
    \item \textbf{Latent-c2f}: Enhancing \textbf{Latent} with a coarse-to-fine parsing strategy, as described in Section~\ref{subsection:coarse2fine-decoding}.
\end{itemize}
Results using pseudo rightward structures and annotated structures are provided in Appendix~\ref{appendix:additional-models}.

\paragraph{Hyper-parameters.}
All models utilize the ``bert-base-chinese''\footnote{\href{https://huggingface.co/bert-base-chinese}{https://huggingface.co/bert-base-chinese}} as the encoder to obtain contextual representations.
For word-level models, word representations are derived by averaging the corresponding character representations. 
The configuration of the scoring layer adheres to \citet{zhang-etal-2020-efficient}.
Refer to Appendix~\ref{appendix:train-detail} for detailed hyper-parameter settings and optimization procedures.
All results are averaged over four runs with different random seeds.

\subsection{Main Results}

\paragraph{Comparison with the pipeline framework.}
As shown in Table~\ref{table:joint-parsing}, our latent models (Latent and Latent-c2f) consistently outperform the pipeline model across all metrics, except for LF$_{dep}$ on CTB5 using SD, where Latent-c2f is lower by 0.05\%.
Latent-c2f achieves absolute improvements of 0.27\% and 0.37\% in LF$_{dep}$ score on CTB6 across two dependency representations.
Similar improvements are observed on CTB5 and CTB7.
The results demonstrate the efficacy of our proposed latent parsing method in mitigating the error propagation problem.

\paragraph{Comparison with previous joint models.}
Table~\ref{table:joint-parsing} also compares our method against previous joint models.
The majority of prior models rely on traditional discrete features or static embeddings, resulting in performance lag compared to our latent models.
The exception is \citet{yan-etal-2020-graph}, which utilizes pre-trained BERT. 
Nevertheless, our latent models achieve substantial improvements, e.g., a 1.52\% increase in LF$_{dep}$ on CTB7.

Notably, Leftward can be considered a reimplementation of \citet{yan-etal-2020-graph}, employing the same network architecture and hyper-parameter settings as our latent models.
In comparison, Latent achieves comparable parsing performance and Latent-c2f achieves better parsing performance.

\begin{table}[tb]
    \setlength{\tabcolsep}{4pt}
    
    \centering
    \begin{tabular}{lcccc}
    \toprule
    \multirow{3}{*}[0.75ex]{Model} & \multicolumn{2}{c}{CTB5} & \multicolumn{2}{c}{CTB7}  \\\cmidrule(lr){2-3}\cmidrule(lr){4-5}
    & UAS & LAS & UAS & LAS  \\
    \midrule
    \multicolumn{5}{c}{\emph{w/ head-finding rules of SD} } \\[2pt]
    TreeCRF & 92.83  & 90.14  & \textbf{90.14} & \textbf{85.89} \\
    \xmidrule[5]
    Leftward &92.69 &  89.91 & 89.08 & 84.77 \\
    Latent (\emph{Ours}) &  \textbf{92.99} &  \textbf{90.19}  & 89.29 &84.99   \\
    Latent-c2f (\emph{Ours}) & 92.84  & 89.99  & 89.69  & 85.45 \\
    \midrule
    \multicolumn{5}{c}{\emph{w/ head-finding rules of Z\&C} } \\[2pt]
    TreeCRF    &  93.95 & 92.90   & \textbf{90.52} & \textbf{89.16} \\
    \xmidrule[5]
    Leftward &  93.77 & 92.70 & 89.67 & 88.26 \\
    Latent (\emph{Ours}) & \textbf{93.96}  & \textbf{93.00}  & 89.86 & 88.46 \\
    Latent-c2f (\emph{Ours}) &  93.88 &  92.89 & 90.06  & 88.70 \\
    \bottomrule
    \end{tabular}
    \caption{Results using gold-standard segmentation on CTB5 and CTB7 test sets.
            Best results are in bold.
    }\label{table:word-level-parsing-results}
\end{table}

\paragraph{Parsing with gold-standard segmentation.}
To isolate the impact of word segmentation errors on parsing performance, we also conduct experiments using gold-standard segmentation with the constrained Eisner algorithm, employing attachment score metrics (UAS and LAS).

As shown in Table~\ref{table:word-level-parsing-results}, character-level models lag behind the word-level model (TreeCRF) by a significant margin, except for Latent on CTB5.\footnote{This discrepancy may be attributed to the utilization of word-level information. Unlike word-level models that can directly utilize word representations, character-level models are merely aware of word boundaries.}
Among character-level models, Latent-c2f significantly enhances the performance of Latent on CTB7 and two latent models consistently outperform Leftward. This suggests that our latent models possess a superior ability in identifying head characters of words, and \emph{enforcing the rightmost character as the word head may not be the best practice}.

\subsection{Analysis}

\paragraph{Impact of proposed constraints.}

\begin{table}[tb]
    \setlength{\tabcolsep}{4pt}
    
    \centering
    \begin{tabular}{lcccc}
    \toprule
    \multirow{3}{*}[0.75ex]{Model} & \multicolumn{4}{c}{CTB7} \\\cmidrule{2-5}
    & F1$_{seg}$ & UF$_{dep}$ &  LF$_{dep}$ & CM \\
    \midrule
    \multicolumn{5}{c}{\emph{w/ head-finding rules of SD} } \\[2pt]
    Latent-c2f        & \textbf{97.12}     &  \textbf{85.59}    &  \textbf{81.87} & \textbf{29.22} \\
    - single-root     & 96.89  & 85.31   &   81.57  & 28.16  \\
    - root-as-head    &  97.07 & 85.52  &  81.79   & 28.59 \\
    - both constraints    & 96.80    & 85.21   & 81.48   & 27.26     \\
    \bottomrule
    \end{tabular}
    \caption{Ablation study on CTB7 dev set. ``CM'': Complete match of labeled dependency trees.}
    \label{table:ablation-study}
\end{table}

Ablation studies are conducted to investigate the individual and combined effects of the single-root subtree and root-as-head constraints on the constrained Inside algorithm. 
Complete match (CM) scores for the entire dependency tree are also provided.
The removal of both constraints, as shown in Table~\ref{table:ablation-study}, results in the lowest LF$_{dep}$ score.
The individual application of each constraint is less effective than using both constraints together.
Notably, the absence of the single-root subtree constraint leads to a more significant decline in performance. 
This is justified by the fact that the single-root subtree constraint minimizes the segmentation of words into disjoint parts.
The application of the root-as-head constraint alone offers a modest 0.08\% improvement in LF$_{dep}$ but leads to a substantial 0.63\% increase in CM.
The results indicate that \emph{an accurate representation of intra-word structures and their syntactic relationships is beneficial for parsing performance and tree completeness.}

\paragraph{Distribution of predicted intra-word structures.}
\begin{table}[tb]
    \setlength{\tabcolsep}{4pt}
    
    \centering
    \begin{tabular}{lrrrrr}
    \toprule
    \multirow{3}{*}[1ex]{Structure} & \multicolumn{2}{c}{Latent} & \multicolumn{2}{c}{Latent-c2f} & \multirow{3}{*}[1ex]{Annt.} \\\cmidrule(lr){2-3}\cmidrule(lr){4-5}
    & \multicolumn{1}{c}{SD} & \multicolumn{1}{c}{Z\&C} & \multicolumn{1}{c}{SD} & \multicolumn{1}{c}{Z\&C}  & \\
    \midrule
    \tikz[]{\structbr} &  99.52   &  99.70 &  49.32 & 50.26 & 48.07 \\ 
    \tikz[]{\structra} &  0.48  &  0.30 & 50.68 & 49.74 & 51.93  \\
    \xmidrule[6]
    \tikz[]{\structbcr} &  91.34  &  91.32 &  41.04 & 42.39 & 34.67 \\
    \tikz[]{\structcar} &   0.02  &  0.02 & 4.67 & 1.64 & 34.20 \\
    \tikz[]{\structraa} &  0.01   &  0.00 & 40.20 & 37.64 &  1.87 \\
    \xmidrule[6]
    \tikz[]{\structbcdr} &  54.06   & 55.34  & 10.07 & 9.28 & 7.24 \\
    \tikz[]{\structraaa} &  0.00   & 0.00  & 21.33 & 30.83 & 0.07 \\
    \tikz[]{\structbrdb} &  2.77   & 2.66  & 1.78 & 0.98 & 15.45 \\
    \tikz[]{\structcarc} &  0.00   & 0.00  & 5.27 & 2.48 & 7.20 \\
    \bottomrule
    \end{tabular}
    \caption{Distribution of intra-word structures predicted by our latent models on CTB6 test set. ``Annt.'' denotes annotated structures. Only high-frequency structures are provided. Filled dots represent root characters.}\label{table:word-internal-struct}
\end{table}

A unique feature of our method is its capacity to infer complex intra-word structures.
We assess the distribution of predicted structures by the constrained Eisner algorithm, grouping them by word length to evaluate common patterns.\footnote{The complete match evaluation is presented in Appendix~\ref{appendix:struct-cm}.}
We focus on words of two, three, and four characters, as longer words are infrequent.
A reference distribution of annotated structures by \citet{gong-etal-2021-depth} is also provided.
High-frequency structures are shown in Figure~\ref{table:word-internal-struct}.
A comprehensive overview is available in Table~\ref{table:all-word-internal-struct} in the appendix.

For Latent, a prevalent left-wavy pattern emerges across words of varying lengths. 
Latent-c2f alleviates this leftward bias.
For two-character words, the left-headed and right-headed structures in Latent-c2f are balanced, closely aligning with the annotated ones.
For three- and four-character words, Latent-c2f can predict right-branched structures, which are seldom or never observed in Latent.

The leftward bias in Latent deserves further discussion.
The Latent model, employing the Eisner algorithm, does not distinctly differentiate between intra- and inter-word dependencies. 
Consequently, this conflation unintentionally transfers the arc direction bias from the inter-word dependencies—derived from word-level trees—to the inherently latent intra-word dependencies.
Given that Chinese is a left-branching language, 
CTB exhibits a predominant occurrence of leftward arcs over rightward ones, with a distribution of 60\% on SD and 70\% on Z\&C.
The Latent-c2f model utilizes dual biaffine attention mechanisms for scoring dependencies, which serves to selectively filter arc direction information, thereby mitigating the inherent leftward bias observed in Latent.

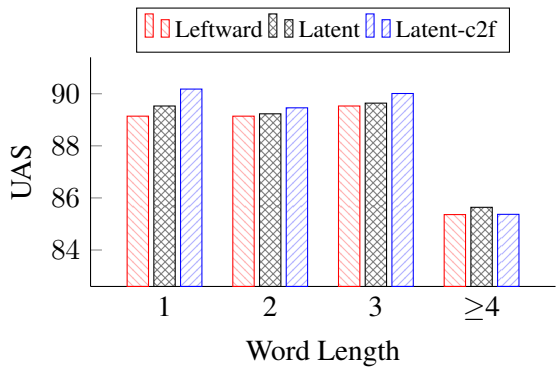
\begin{figure}
    \centering
    \begin{tikzpicture}
        \begin{axis}
            [
                width=\columnwidth,
                height=0.6\columnwidth,
                ybar,
                bar width=8pt,
                enlargelimits=0.05,
                axis x line*=bottom,
                axis y line*=left,
                xlabel={Word Length},
                ylabel={UAS},
                xmin=0.5,
                xmax=4.5,
                ymin=83,
                ymax=91,
                tick align=inside,
                xtick={1,2,3,4},
                xticklabels={1,2,3,$\geq$4},
                xtick style={draw=none},
                legend style={at={(0.5, 1.02)}, anchor=south, font=\small},
                legend columns=-1,
                legend entries={Leftward, Latent, Latent-c2f},
            ]
            \addplot [draw=red, pattern=north west lines, pattern color=red!30!white] coordinates {
                (1,89.14)
                (2,89.14)
                (3,89.53)
                (4,85.36)
            };
            \addplot[draw=black, pattern=crosshatch, pattern color=gray] coordinates {
                (1,89.53)
                (2,89.23)
                (3,89.64)
                (4,85.64)
            };
            \addplot[draw=blue, pattern=north east lines, pattern color=blue!30!white] coordinates {
                (1,90.18)
                (2,89.46)
                (3,90.01)
                (4,85.37)
            };
        \end{axis}
    \end{tikzpicture}
    \caption{The unlabeled attachment score (UAS) for words of different lengths on CTB7 test set using SD.}\label{fig:intra-word-uas}
\end{figure}

\paragraph{Performance across word lengths.}
We further investigate the performance of character-level models across words of different lengths.
The results in Figure~\ref{fig:intra-word-uas} are obtained using gold-standard segmentation.
Latent-c2f exhibits the best performance for words of lengths 1, 2, and 3.
However, for words with a length greater than or equal to 4, Latent-c2f performs worse than Latent, suggesting that coarse-to-fine parsing may not be advantageous for longer words.
Interestingly, the performance difference between Leftward and Latent is marginal for words of length 2 and 3.
This is consistent with the information in Table~\ref{table:word-internal-struct}, where intra-word structures of lengths 2 and 3 primarily exhibit a left-wavy pattern for Latent, nearly identical to Leftward.

\section{Related Work}

\paragraph{Intra-word structure.} 
\citet{zhao-2009-character}  were the first to explore intra-word structures in Chinese through unlabeled dependency forms. 
\citet{li-2011-parsing} and \citet{zhang-etal-2013-chinese} extended this work by introducing constituency trees to depict these structures, which were further refined by \citet{zhang-etal-2014-character} through their conversion of constituency trees into dependency trees. 
\citet{gong-etal-2021-depth} went on to investigate intra-word (labeled) dependencies, positioning the parsing of these structures as a distinct task.

\paragraph{Character-level dependency parsing.}
The area of character-level dependency parsing, especially within the context of Chinese, has undergone significant evolution.
\citet{hatori-etal-2012-incremental} led the initial efforts by introducing a transition-based parser that leveraged pseudo intra-word structures. 
This was followed by \citet{zhang-etal-2014-character}, who integrated annotated intra- and inter-word dependencies. 
Subsequent studies aimed to enhance the transition-based parsers with neural networks \citep{kurita-etal-2017-neural,li2018neural}.
\citet{yan-etal-2020-graph} were the first to adopt the graph-based parsing approach.

\paragraph{Span constraints.}
The dependency structure is closely related to spans (not limited to phrases and words).
\citet{spitkovsky-etal-2010-profiting} demonstrated how naturally annotated spans could be transformed into dependency structures by applying various parsing constraints. 
For transition-based parsers, \citet{nivre-etal-2014-squibs} emphasized the necessity of a single-root subtree over the input spans.
Similarly, \citet{zhang-etal-2022-semantic} framed span-based semantic role labeling as dependency parsing, enforcing semantic arguments corresponding to single-root subtrees.

Character-level Dependency Annotation of Chinese
A Truly Joint Neural Architecture for Segmentation and Parsing

\section{Conclusion}

This paper explores modeling latent intra-word structures for character-level Chinese dependency parsing.
Our approach, underpinned by the constrained Eisner algorithm, ensures the compatibility of constructed character-level trees.
The incorporation of a coarse-to-fine parsing strategy further enhances the effectiveness and rationality of the parsing process.
Our experiments and detailed analyses reveal the following findings:
\begin{itemize}[leftmargin=*]
    \item Our method outperforms not only the pipeline model but also previous joint models in character-level Chinese dependency parsing.
    \item Given gold-standard segmentation, our latent models, especially the coarse-to-fine one, demonstrate superior capability in identifying the head character of a word, suggesting that designating the rightmost character as the head of the word may not be optimal.
    \item The proposed compatibility constraints can improve both parsing accuracy and the completeness of tree structures.
    \item The intra-word structures predicted by the latent model tend to exhibit a left-wavy shape. The coarse-to-fine strategy alleviates the leftward bias and produces structures more aligned with manually annotated ones.
\end{itemize}

\section*{Limitations}\label{section:limitations}

\paragraph{Projectivity.}
Our method treats intra-word structures as latent, offering a flexible and rich representation of internal word structures.
However, it operates within the confines of projective parsing due to the inherent nature of the Eisner algorithm. 
This constraint might limit the applicability of the model in accurately parsing non-projective trees.

\paragraph{Computational Efficiency.}
The introduction of constraints into the Eisner algorithm undoubtedly increases its complexity. 
Although auxiliary tensors and GPU utilization help mitigate the additional time burden, computational efficiency remains a concern, particularly as the necessity to calculate inside scores twice doubles the training duration.
Moreover, the incorporation of a coarse-to-fine strategy, while beneficial for parsing accuracy, further compounds the computational demands.

\section*{Ethics Statement}
We are committed to upholding high ethical standards throughout this paper. 
Our research focuses on Chinese dependency parsing, utilizing the Penn Chinese Treebank (LDC2010T07) for experimental purposes. 
We have obtained the necessary permissions and licenses for the acquisition of the data, and we strictly adhere to the terms of use associated with it. 
Researchers with access to the treebank can replicate our experiments using our provided code.
Moreover, the annotated intra-word structures used for analysis are openly accessible and do not impose any acquisition or usage requirements.
We believe that the utilization of these datasets will not compromise the confidentiality or integrity of individuals, nor will it contain offensive content.
Additionally, given that our work primarily explores syntactic methodologies, we do not foresee any potential risks associated with our research. 

\section*{Acknowledgements}

We thank all the anonymous reviewers for their valuable comments. We also thank Houquan Zhou and Cheng Gong for their helpful suggestions during the paper writing process. This work was supported by National Natural Science Foundation of China (Grant No. 62176173 and 62336006), and a
Project Funded by the Priority Academic Program Development (PAPD) of Jiangsu Higher Education Institutions.

\bibliography{anthology,custom}

\appendix
\section{Implementation Details}\label{appendix:implementation}

\subsection{Char-Tree to Word-Tree Recovery}\label{appendix:tree-recovery}

After predicting an optimal character-level tree, a word-level tree can be recovered from it. 
The first step is to identify all subtrees corresponding to words, which must satisfy two conditions:
\begin{enumerate*}[label=(\arabic*)]
    \item contain only intra-word dependency arcs (indicated by an $\textsc{Intra}$ label); 
    \item be linked by an inter-word dependency arc (indicated by common syntactic labels).
\end{enumerate*}
Next, these subtrees are collapsed into words.
Finally, the character-level inter-word arcs are revived into word-level arcs.

\subsection{Loss Function}\label{appendix:loss-function}
To calculate the label loss, we need to enumerate each arc in each tree in the forest, which is exponential in the worst case.
Inspired by \citet{zhang-etal-2022-semantic}, we find this enumeration can be integrated into the computation of the tree loss.

First, we define the probability of assigning the labels to all arcs in the unlabeled tree $\boldsymbol{y}$ as:
\begin{equation}
    p(\boldsymbol{r}|\boldsymbol{x},\boldsymbol{y}) = \prod_{(i,j) \in \boldsymbol{y}} p(l | i,j)
\end{equation}
where $\boldsymbol{r}$ is the set of labels for all arcs in $\boldsymbol{y}$.

Then, we define the probability of the labeled tree $\boldsymbol{t}$ of a given sentence $\boldsymbol{x}$ as:
\begin{equation}
    p(\boldsymbol{t}|\boldsymbol{x}) = p(\boldsymbol{y}|\boldsymbol{x}) \cdot  p(\boldsymbol{r}|\boldsymbol{x},\boldsymbol{y})
\end{equation}

Finally, the loss function is defined as the negative log-likelihood of the labeled forest $\mathcal{T}$:
\begin{equation}
\begin{aligned}
    \mathbf{\emph{L}}(\boldsymbol{x}) &= -\log p(\mathcal{T}|\boldsymbol{x}) \\
    p(\mathcal{T}|\boldsymbol{x}) &= \sum_{\boldsymbol{t} \in \mathcal{T}} p(\boldsymbol{t}|\boldsymbol{x}) \\
    &=\frac{\sum_{\boldsymbol{y} \in \mathcal{F}} e^{s(\boldsymbol{x},\boldsymbol{y})} \cdot p(\boldsymbol{r}|\boldsymbol{x},\boldsymbol{y})}{\boldsymbol{Z(x)}} \\
    &= \frac{\sum_{\boldsymbol{y} \in \mathcal{F}} \prod_{(i,j) \in \boldsymbol{y}} e^{s(i,j) + \log p(l|i,j)}}{\boldsymbol{Z(x)}}
\end{aligned}
\end{equation}

By adding the log probability of labels to the arc scores, the label loss is naturally integrated into the tree loss via the constrained Inside algorithm.
\begin{table}[tb!]
    \centering
    \begin{tabular}{lrrr}
    \toprule
    Dataset & Train & Dev & Test  \\
    \midrule
    CTB5  & 18,104 & 352 & 348 \\
    CTB6  & 23,420 & 2,079 & 2,796 \\
    CTB7  & 31,112 & 10,043 & 10,292 \\
    \bottomrule
    \end{tabular}
    \caption{Data statistics. We present the number of sentences in the training, development, and test sets.
    }\label{table:data-statistic}
\end{table}

\begin{table}[tb]
    \setlength{\tabcolsep}{4pt}
    
    \centering
    \begin{tabular}{lrrrrr}
    \toprule
    \multirow{3}{*}[1ex]{Structure} & \multicolumn{2}{c}{Latent} & \multicolumn{2}{c}{Latent-c2f} & \multirow{3}{*}[1ex]{Annt.} \\\cmidrule(lr){2-3}\cmidrule(lr){4-5}
    & \multicolumn{1}{c}{SD} & \multicolumn{1}{c}{Z\&C} & \multicolumn{1}{c}{SD} & \multicolumn{1}{c}{Z\&C}  & \\
    \midrule
    \tikz[]{\structbr} &  99.52   &  99.70 &  49.32 & 50.26 & 48.07 \\ 
    \tikz[]{\structra} &  0.48  &  0.30 & 50.68 & 49.74 & 51.93  \\
    \xmidrule[6]
    \tikz[]{\structbcr} &  91.34  &  91.32 &  41.04 & 42.39 & 34.67 \\
    \tikz[]{\structcar} &   0.02  &  0.02 & 4.67 & 1.64 & 34.20 \\
    \tikz[]{\structccr} &  7.03   & 8.10  & 8.15 & 7.31 & 5.78 \\
    \tikz[]{\structraa} &  0.01   &  0.00 & 40.20 & 37.64 &  1.87 \\
    \tikz[]{\structrab} &  0.02   & 0.01  & 2.96 & 9.14 & 7.02 \\
    \tikz[]{\structbrb} &   0.96  &  0.19 & 2.37 & 1.60 & 15.30 \\
    \xmidrule[6]
    \tikz[]{\structbcdr} &  54.06   & 55.34  & 10.07 & 9.28 & 7.24 \\
    \tikz[]{\structbddr} &  12.37   & 22.33  & 9.28 & 12.08 & 9.39 \\
    \tikz[]{\structdddr} &  0.44   & 1.98  & 18.72 & 14.79 & 0.57 \\
    \tikz[]{\structraaa} &  0.00   & 0.00  & 21.33 & 30.83 & 0.07 \\
    \tikz[]{\structraac} &  0.04   & 0.00  & 6.19 & 3.92 & 11.94 \\
    \tikz[]{\structbrdb} &  2.77   & 2.66  & 1.78 & 0.98 & 15.45 \\
    \tikz[]{\structcarc} &  0.00   & 0.00  & 5.27 & 2.48 & 7.20 \\
    \tikz[]{\structbcrc} &  0.32   & 0.20  & 0.20 & 0.00 & 7.13 \\
    \bottomrule
    \end{tabular}
    \caption{Distribution of intra-word structures predicted by our latent models on the CTB6 test set. ``Annt.'' denotes annotated structures. Filled dots represent root characters.}\label{table:all-word-internal-struct}
\end{table}

\subsection{Coarse-to-fine Scoring}\label{appendix:c2f-scoring}
To score a dependency arc $i \rightarrow j$, we first feed the output from encoder $\mathbf{h}$ into two MLPs to obtain the representations of character as head and modifier.
Then, to distinguish the intra-word and inter-word roles, the arc is scored by two different biaffine layers.
\begin{equation}
     \begin{aligned}
        \mathbf{h}^{(arc-head)}_i & = \mathbf{MLP}^{(arc-head)}(\mathbf{h}_i) \\
        \mathbf{h}^{(arc-mod)}_j & = \mathbf{MLP}^{(arc-mod)}(\mathbf{h}_j) \\
        s^{(intra)}(i,j) & = \mathbf{h}^{(arc-head)}_iW^{(intra)}\mathbf{h}^{(arc-mod)}_j \\
        s^{(inter)}(i,j) & = \mathbf{h}^{(arc-head)}_iW^{(inter)}\mathbf{h}^{(arc-mod)}_j
    \end{aligned}
\end{equation}
\subsection{Hyper-parameter Details}\label{appendix:train-detail}
We utilize the default parameter configurations for pre-trained BERT and directly fine-tune the entire model.
The configuration of the scoring layer adheres to \citet{zhang-etal-2020-efficient}.
We employ AdamW \citep{DBLP:conf/iclr/LoshchilovH19} for parameter optimization with $\beta_1 = 0.9$, $\beta_2 = 0.9$, $\epsilon = 1 \times 10^{-12}$, and weight decay of $0$.
The learning rate is set to $5 \times 10^{-5}$ for the encoder and $1 \times 10^{-3}$ for the scorer.
The dropout rate is set to $0.1$ for the encoder and $0.33$ for the scorer.
We train the model for 10 epochs with 1,000 tokens per batch.

\begin{table*}[tb]
    \centering
    \begin{tabular}{lccccccccc}
    \toprule
    \multirow{3}{*}[0.75ex]{Model} & \multicolumn{3}{c}{CTB5} & \multicolumn{3}{c}{CTB6} & \multicolumn{3}{c}{CTB7} \\\cmidrule(lr){2-4}\cmidrule(lr){5-7}\cmidrule(lr){8-10}
    & F1$_{seg}$ & UF$_{dep}$ & LF$_{dep}$ & F1$_{seg}$ & UF$_{dep}$ & LF$_{dep}$ & F1$_{seg}$ & UF$_{dep}$ & LF$_{dep}$\\
    \midrule
    \multicolumn{10}{c}{\emph{w/ head-finding rules of SD} } \\[2pt]
    Pipeline & 98.72 & 90.93 & 88.39 & 97.23 & 87.09 & 83.86 & 97.16 & 85.77 & 82.00 \\
    Annotated & 98.68 & 90.74 & 88.05 & 97.30 & 87.10 & 83.87 & 97.17 & 85.70 & 81.95 \\
    Leftward & 98.76 & 90.91 & 88.37 & 97.30 & 87.21 & 84.04 & 97.22 & 85.85 & 82.17 \\
    Rightward & 98.76 & 90.83 & 88.24 & \textbf{97.35} & \textbf{87.34} & \textbf{84.12} & \textbf{97.23} & 85.89 & \textbf{82.23} \\
    Latent (\emph{Ours}) & 98.76 & \textbf{91.06} & \textbf{88.49} & 97.28 & 87.22 & 84.03 & 97.17 & 85.74 & 82.04 \\
    Latent-c2f (\emph{Ours}) & \textbf{98.79} & 90.95 & 88.34 & 97.33 & 87.30 & \textbf{84.12} & 97.22 & \textbf{85.90} & \textbf{82.23} \\
    \midrule
    \multicolumn{10}{c}{\emph{w/ head-finding rules of Z\&C} } \\[2pt]
    Pipeline & 98.72 & 92.00 & 91.04 & 97.23 & 87.71 & 86.77 & 97.16 & 86.39 & 85.19 \\
    Annotated & 98.66 & 91.85 & 90.92 & 97.34 & 87.87 & 86.90 & 97.23 & 86.35 & 85.17 \\
    Leftward & 98.67 & 91.98 & 91.03 & \textbf{97.39} & 88.02 & 87.06 & \textbf{97.26} & 86.48 & 85.30 \\
    Rightward & 98.72 & 91.65 & 90.71 & 97.33 & 87.84 & 86.90 & \textbf{97.26} & 86.46 & 85.28 \\
    Latent (\emph{Ours}) & 98.74 & \textbf{92.16} & \textbf{91.25} & 97.37 & 87.99 & 87.06 & 97.22 & 86.50 & 85.33 \\
    Latent-c2f (\emph{Ours}) & \textbf{98.77} & 92.03 & 91.08 & 97.37 & \textbf{88.10} & \textbf{87.14} & \textbf{97.26} & \textbf{86.55} & \textbf{85.37} \\
    \bottomrule
    \end{tabular}
    
    \caption{Results on CTB5, CTB6, and CTB7 test sets. 
            The best results are in bold.}\label{table:all-result}
\end{table*}

\begin{table}[tb]
    \centering
    \begin{tabular}{lcc}
    \toprule
    Model & CM & CM$_\text{M-1}$ \\
    \midrule
    Latent (SD) & 42.86  & 44.20 \\
    Latent (Z\&C) & 42.77  & 44.11 \\
    Latent-c2f (SD) & \textbf{44.26} & \textbf{85.00} \\
    Latent-c2f (Z\&C) & 42.41 & 84.36 \\

    \bottomrule
    \end{tabular}
    \caption{Complete match (CM) of intra-word structures on CTB6 test set.}\label{table:struct-cm}
\end{table}

\section{Supplementary Results}\label{appendix:supp_result}

\subsection{Additional Models}\label{appendix:additional-models}
Two additional models are included in the comparison, employing different strategies to handle the internal structures of words:
\begin{itemize}[leftmargin=*]
    \item \textbf{Rightward:} A model uses pseudo intra-word structures in a right-wavy pattern, which is similar to the leftward pattern but in the opposite direction.
    \item \textbf{Annotated:} A model uses annotated intra-word structures by \citet{gong-etal-2021-depth}. If no annotated structure is available for a word, the latent structure is employed.
\end{itemize}

The results are presented in Table~\ref{table:all-result}.
Rightward achieves performance similar to Leftward.
Specifically, it performed slightly better on SD but slightly worse on Z\&C.
Surprisingly, Annotated only achieves comparable performance to Pipeline.
Comparing Annotated and Latent, the use of annotated structures does not improve performance and even degrades it.
This finding is consistent with \citet{wu2021deep}, who observed that using annotated structures by \citet{zhang-etal-2014-character} is detrimental to neural dependency parsers.
Two points can be concluded from the results:
\begin{itemize}[leftmargin=*]
    \item Both leftward and rightward intra-word structures are effective for the joint CWS and dependency parsing task.
    \item The usefulness of annotated structures in the deep learning era is questionable and deserves further investigation.
\end{itemize}

\subsection{Complete Match of Structures}\label{appendix:struct-cm}
In addition to investigating the distribution of intra-word structures, we utilize the complete match (CM) metric to evaluate the performance of our latent models in predicting intra-word structures.
The complete match measures the percentage of words with correct whole structures.
Here, we refer to the intra-word structures annotated by \citet{gong-etal-2021-depth} as the gold standard.
We calculate the average of the results from four seed models.
Additionally, since no gold-standard structures are employed during training, the evaluation can be regarded as unsupervised.
Following studies on unsupervised POS tagging \citep{johnson-2007-doesnt,tran-etal-2016-unsupervised}, we employ a many-to-one (M-1) mapping to align the predicted structures with the gold standard.
Specifically, if any predicted structure by a seed model matches the gold standard, it is considered a complete match.
The results are shown in Table~\ref{table:struct-cm}.
Compared to Latent, Latent-c2f achieves a similar CM score but higher M-1 mapping results.
This is because Latent-c2f favors leftward arcs in some seed models and rightward arcs in others.
When employing a many-to-one mapping, more structures predicted by Latent-c2f align with their gold-standard counterparts.

\begin{algorithm*}[tb]
    \caption{Coarse-to-fine Eisner Algorithm.}\label{alg:c2f-eisner}
    {
        \begin{algorithmic}[1]
        \State \textbf{Input}: intra-word arc scores $\hat{s}(i,j)$ and inter-word arc scores $s(i,j)$ 
        \State \textbf{Define}: $\hat{I},I,\hat{C},C \in \mathbf{R}^{n \times n}$ 
        \Comment{The hat symbol denotes an intra-word span}
        \State \textbf{Initialize}: $\hat{C}_{i \rightarrow i} = 0, C_{i \rightarrow i} = -\infty, 1 \leq i \leq n$
        \For {$w = 1, \ldots, \ n$}
            \For {$i = 1, \ldots, \ n - w$}
                \State {$j = i + w$}
                \State $\begin{aligned}
                    \hat{I}_{i \rightarrow j} = \max_{i \le k < j}(\hat{s}(i,j) + \hat{C}_{i \rightarrow k} + \hat{C}_{k+1 \leftarrow j})
                \end{aligned}$ 
                \State $\begin{aligned}
                    I_{i \rightarrow j} = \max_{i \le k < j}(& s(i,j) + \hat{C}_{i \rightarrow k} + \hat{C}_{k+1 \leftarrow j}, s(i,j) + \hat{C}_{i \rightarrow k} + C_{k+1 \leftarrow j}, \\
                                               & s(i,j) + C_{i \rightarrow k} + \hat{C}_{k+1 \leftarrow j}, s(i,j) + C_{i \rightarrow k} + C_{k+1 \leftarrow j})
                \end{aligned}$ 
                \State $\begin{aligned}
                    \hat{I}_{i \leftarrow j} = \max_{i \le k < j}(\hat{s}(j,i) + \hat{C}_{i \rightarrow k} + \hat{C}_{k+1 \leftarrow j})
                \end{aligned}$ 
                \State $\begin{aligned}
                    I_{i \leftarrow j} = \max_{i \le k < j}(& s(j,i) + \hat{C}_{i \rightarrow k} + \hat{C}_{k+1 \leftarrow j}, s(j,i) + \hat{C}_{i \rightarrow k} + C_{k+1 \leftarrow j}, \\
                                               & s(j,i) + C_{i \rightarrow k} + \hat{C}_{k+1 \leftarrow j}, s(j,i) + C_{i \rightarrow k} + C_{k+1 \leftarrow j})
                \end{aligned}$ 
                \State $\begin{aligned}
                    \hat{C}_{i \rightarrow j} = \max_{i < k \le j}(\hat{I}_{i \rightarrow k} + \hat{C}_{k \rightarrow j})
                \end{aligned}$
                \State $\begin{aligned}
                    C_{i \rightarrow j} = \max_{i < k \le j}(\hat{I}_{i \rightarrow k} + C_{k \rightarrow j}, I_{i \rightarrow k} + \hat{C}_{k \rightarrow j}, I_{i \rightarrow k} + C_{k \rightarrow j}) \\
                \end{aligned}$
                \State $\begin{aligned}
                    \hat{C}_{i \leftarrow j} = \max_{i \le k < j}(\hat{C}_{i \leftarrow k} + \hat{I}_{k \leftarrow j})
                \end{aligned}$
                \State $\begin{aligned}
                    C_{i \leftarrow j} = \max_{i \le k < j}(C_{i \leftarrow k} + \hat{I}_{k \leftarrow j}, \hat{C}_{i \leftarrow k} + I_{k \leftarrow j}, C_{i \leftarrow k} + I_{k \leftarrow j}) \\
                \end{aligned}$
            \EndFor
        \EndFor
        \State \Return  $C_{1 \rightarrow n}$
        \end{algorithmic}
    }
\end{algorithm*}

\end{CJK}
\end{document}